%% file: _arxiv.tex
\documentclass{article}

\usepackage[preprint]{neurips_2023}
\usepackage[breaklinks,colorlinks=true, citecolor=cite, linkcolor=ref]{hyperref}

\input{common}

\title{Synthetic Shifts to Initial Seed Vector Exposes \\the Brittle Nature of Latent-Based Diffusion Models}

\author{%
    Mao Po-Yuan$^*$ \quad Shashank Kotyan$^*$ \quad Tham Yik Foong \quad Danilo Vasconcellos Vargas \\
    Laboratory of Intelligent Systems \\
    Kyushu University, Fukuoka, Japan \\
}

\begin{document}
    \maketitle
    \def\thefootnote{*}\footnotetext{These authors contributed equally to this work}\def\thefootnote{\arabic{footnote}}

    \input{content}

\end{document}

%% file: common.tex
\usepackage[T1]{fontenc}   
\usepackage[utf8]{inputenc} 
\usepackage{algorithm, algorithmicx, algpseudocode}
\usepackage{amsmath, amsfonts, amssymb}
\usepackage{array}
\usepackage{booktabs}
\usepackage{collcell}
\usepackage{datatool}
\usepackage{enumitem}
\usepackage{environ}
\usepackage{etoolbox}
\usepackage{flushend}
\usepackage{graphicx}
\usepackage{lipsum}
\usepackage{microtype}
\usepackage{multirow}
\usepackage{nicefrac}
\usepackage{printlen}
\usepackage{soul}
\usepackage{svg}
\usepackage{url}
\usepackage{wrapfig}
\usepackage{xcolor,colortbl}
\usepackage{xstring}
\usepackage{multicol}

\definecolor{Gray}{gray}{0.85}
\definecolor{cite}{HTML}{53769A}
\definecolor{ref}{HTML}{379030}
\definecolor{lightcornflowerblue}{rgb}{0.6, 0.81, 0.93}
\definecolor{lightkhaki}{rgb}{0.94, 0.9, 0.55}
\definecolor{lightmauve}{rgb}{0.86, 0.82, 1.0}
\definecolor{lightgreen}{rgb}{0.56, 0.93, 0.56}

\definecolor{royalpurple}{RGB}{207,199,216}
\definecolor{forestgreen}{RGB}{202,225,204}

\definecolor{PatternA}{RGB}{180, 22,  0   }
\definecolor{PatternC}{RGB}{23,  77,  127 }
\definecolor{PatternB}{RGB}{55,  144, 48  }

\newenvironment{aequation}
{\begin{equation} \begin{aligned}}
{\end{aligned} \end{equation}}

\newenvironment{afigure}
{\begin{figure} \centering}
{\end{figure}}

\newcommand{\StableDiffOld}{Stable Diffusion v1.5}
\newcommand{\StableDiffNew}{Stable Diffusion v2.1}
\newcommand{\StableDiff}{Stable Diffusion~\citep{stable_dif}}
\newcommand{\SD}{Stable Diffusion} 
\newcommand{\GLIDE}{GLIDE~\citep{DBLP:conf/icml/NicholDRSMMSC22}} 
\newcommand{\Glide}{GLIDE}

\newcommand{\RS}{Random Shift}
\newcommand{\MS}{Mean Shift}
\newcommand{\SDS}{Standard Deviation Shift} 
\newcommand{\MixS}{Mixed Shift}
\newcommand{\AS}{Arrangement Shift}

\newcolumntype{H}{>{\setbox0=\hbox\bgroup}c<{\egroup}@{}}

\newcommand{\com}[1]{\iffalse~#1~\fi}%
\newcommand{\x}[1]{{\underline{#1}}}%

\algrenewcommand{\algorithmiccomment}[1]{\hfill$\blacktriangleright$ #1}

\newcolumntype{X}{>{\columncolor{lightcornflowerblue}}c}
\newcolumntype{Y}{>{\columncolor{lightkhaki}}c}
\newcolumntype{Z}{>{\columncolor{lightmauve}}c}
\newcolumntype{P}{>{\columncolor{lightgreen}}c}

\newcolumntype{+}{>{\global\let\currentrowstyle\relax}}
\newcolumntype{^}{>{\currentrowstyle}}

\newcommand{\noimage}{%
  \setlength{\fboxsep}{-\fboxrule}%
  \fbox{\phantom{\rule{100pt}{100pt}}File missing\phantom{\rule{100pt}{100pt}}}
}
\let\includegraphicsoriginal\includegraphics%
\renewcommand{\includegraphics}[2][width=\textwidth]{\IfFileExists{#2}{\includegraphicsoriginal[#1]{#2}}{\noimage}}

\emergencystretch=\maxdimen%
\everypar{\looseness=-1 }

\newcounter{descriptcount}

\newcounter{CurrentRow}
\newcounter{CurrentColumn}
\newtoggle{DoneWithFirstRow}

\newcommand*{\FirstColumn}[1]{%
    \IfEq{\arabic{CurrentColumn}}{0}{%
        \global\togglefalse{DoneWithFirstRow}%
        \setcounter{CurrentRow}{1}
    }{%
        \global\toggletrue{DoneWithFirstRow}%
        \stepcounter{CurrentRow}%
    }%
    \setcounter{CurrentColumn}{0}%
    \NewData{#1}%
}
\newcommand*{\NewData}[1]{%
    \dtlexpandnewvalue%
    \stepcounter{CurrentColumn}%
    \iftoggle{DoneWithFirstRow}{%
        \dtlgetrow{TransposedTabularDB}{\arabic{CurrentColumn}}%
        \dtlappendentrytocurrentrow{\Alph{CurrentRow}}{#1}%
        \dtlrecombine%
    }{%
        \DTLnewrow{TransposedTabularDB}%
        \DTLnewdbentry{TransposedTabularDB}{\Alph{CurrentRow}}{#1}%
    }%
}%
\newcolumntype{+}{>{\global\let\currentrowstyle\relax}}
\newcolumntype{^}{>{\currentrowstyle}}
\newcolumntype{F}{>{\collectcell\FirstColumn}c<{\endcollectcell}}
\newcolumntype{C}{>{\collectcell\NewData}{c}<{\endcollectcell}}

\newtoggle{EncounteredDataRow}

\newsavebox{\TempBox}
\DTLnewdb{TransposedTabularDB}

\NewEnviron{Ttabular}[1]{%
    \setcounter{CurrentColumn}{0}%
    \global\togglefalse{EncounteredDataRow}%
    \savebox{\TempBox}{%
        \begin{tabular}{FCCCCCC}
            \BODY%
        \end{tabular}%
    }%
    \begin{tabular}{#1}\toprule%
    \DTLforeach*{TransposedTabularDB}{\Aa=A, \Ba=B, \Ca=C}{%
      \DTLiffirstrow{}{\\\midrule}%
        \Aa&\Ba&\Ca%
    }\\\bottomrule%
    \end{tabular}%
}%

\newcommand{\figtop}{{\em (Top)}}
\newcommand{\figbottom}{{\em (Bottom)}}

\def\Tableref#1{Table~\ref{#1}}

\def\Figref#1{Figure~\ref{#1}}

\def\Figsref#1#2{Figures~\ref{#1}~-~\ref{#2}}
\def\Twofigref#1#2{Figures~\ref{#1}~and~\ref{#2}}

\def\Secref#1{Section~\ref{#1}}

\def\eqref#1{equation~\ref{#1}}
\def\Eqref#1{Equation~\ref{#1}}

\usepackage{mathabx,graphicx}

\def\1{\bm{1}}

\newcommand{\R}{\mathbb{R}}

\DeclareMathAlphabet{\mathsfit}{\encodingdefault}{\sfdefault}{m}{sl}
\SetMathAlphabet{\mathsfit}{bold}{\encodingdefault}{\sfdefault}{bx}{n}

\def\gN{{\mathcal{N}}}



%% file: content.tex
\begin{abstract}
Recent advances in Conditional Diffusion Models have led to substantial capabilities in various domains. 
However, understanding the impact of variations in the initial seed vector remains an underexplored area of concern. 
Particularly, latent-based diffusion models display inconsistencies in image generation under standard conditions when initialized with suboptimal initial seed vectors. 
To understand the impact of the initial seed vector on generated samples, we propose a reliability evaluation framework that evaluates the generated samples of a diffusion model when the initial seed vector is subjected to various synthetic shifts.
Our results indicate that slight manipulations to the initial seed vector of the state-of-the-art \StableDiff~can lead to significant disturbances in the generated samples, consequently creating images without the effect of conditioning variables.
In contrast, \GLIDE~stands out in generating reliable samples even when the initial seed vector is transformed.
Thus, our study sheds light on the importance of the selection and the impact of the initial seed vector in the latent-based diffusion model.
\end{abstract}

\section{Introduction}

\begin{afigure}
    \includegraphics[width=0.5\columnwidth]{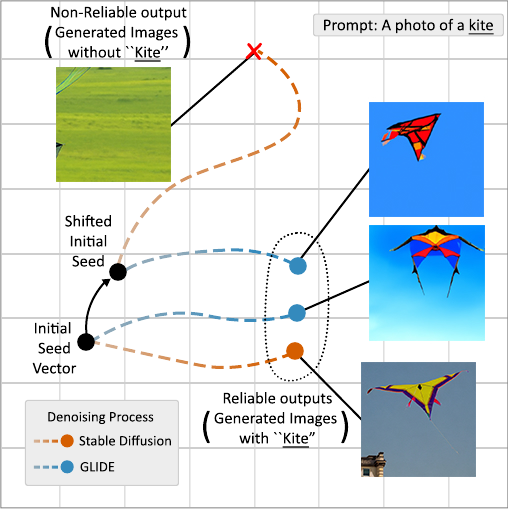}
    \caption{
    \textbf{Illustration of performance of \StableDiff~compared to \GLIDE~in the Presence of Synthetic Shifts to the Initial Seed Vector.}
    This figure illustrates the trajectories of the diffusion process in pixel space for both \StableDiffNew~and \Glide~models when subjected to seed vector shifts.  
    We notice that with a slight shift to the initial seed vector, the generated image by \SD~diverges towards a non-reliable output. 
    On the other hand, \Glide~consistently generates reliable outputs, demonstrating robustness to shift in the initial seed vector. 
    }
    \label{fig:illustration}
\end{afigure}

\begin{figure*}[!ht] 
\centering
    \includegraphics[width=\textwidth]{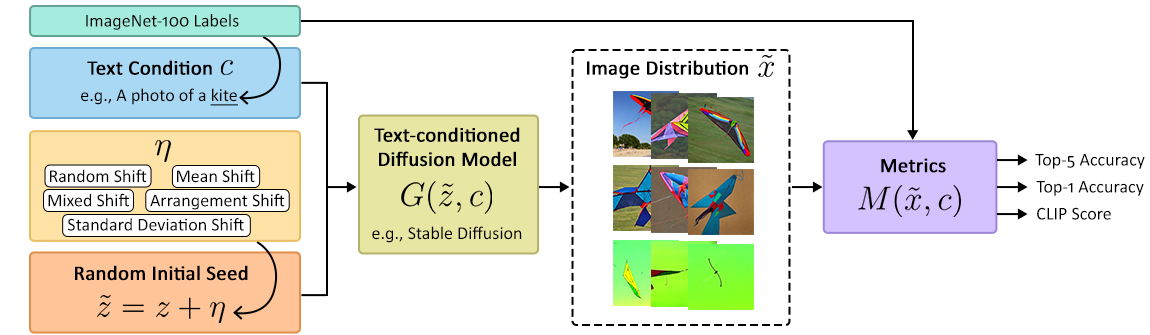}
    \caption{
    \textbf{Illustration of the Reliability Evaluation Framework.} 
    Prompts are constructed by filling the sentence "A photo of a \x{\quad \quad}" with ImageNet100 \citep{ImageNet100} labels, and the random initial seed vector $z$ is transformed by $\eta$ to form a shifted initial seed vector $\bar{z}$. Subsequently, the test-conditioned diffusion model generates an image distribution $\tilde{x}$ based on these inputs.
    The metrics $M$ then evaluate the generated image distribution with the known label used in the conditioning variable. 
    }
\label{fig:framework}
\end{figure*}
In recent years, diffusion models have risen to the forefront as state-of-the-art instruments for content creation and the precision generation of high-quality synthetic data empowered by deep neural networks and extensive datasets~\citep{bao2022analyticdpm}. 
Their influence spans across multiple domains, including images ~\citep{ho2020denoising,dhariwal2021diffusion,ho2022cascaded,ho2022classifier}, audio~\citep{kong2021diffwave,ijcai2022p577,huang2022prodiff,kim2022guided}, texts~\citep{li2022diffusion}, molecules~\citep{xu2022geodiff}, solidifying their status as leading technologies in data synthesis.
Notably, the release of Stable Diffusion~\citep{stable_dif}, an advanced open-source text-based image generation model, has sparked diverse applications and workloads~\citep{lugmayr2022repaint,jeong2023power}. 

\citet{samuel2023all} links the quality of generating a rare concept to the initial seed vector, suggesting the importance of choosing the initial seed vector. 
Using this understanding, researchers want to drive creativity and facilitate high-quality synthetic data generation with the help of manipulation the initial seed vector.



In the \Figref{fig:illustration}, we show that the latent-based diffusion model is more susceptible to generating non-reliable output based on shifts to the initial seed vector, highlighting the importance of selection of the initial seed vector in the latent-based diffusion model. 
Since a comprehensive understanding of the diffusion process to generate samples is still lacking ~\citep{li2023diffusion,li2023alleviating}, we also focus on understanding the intricate landscape of the generated samples by the diffusion models.
To comprehensively evaluate the impact of the initial seed vector in generating samples, 
we meticulously investigate the ability of a diffusion model to adapt from shifts to the initial seed vector. 
These synthetic shifts encompass a spectrum, including \RS, \MS, \SDS, \MixS, and \AS. 
Thus, to assess the generated samples and evaluate their correctness, we propose a model-agnostic reliability evaluation framework, as illustrated in \Figref{fig:framework}. 

~

\noindent
\textbf{Contributions:} 

\begin{description}[style=sameline, leftmargin=*]
\item[Reliability Evaluation Framework:]
We propose a simple framework (\Figref{fig:framework}) for systematically evaluating how diffusion-based models can handle the shift to the initial seed vector.

\item[Brittle nature of \StableDiff:]
Our results show that slight variations to the initial random vector break the \SD~as it creates undesired samples. 

\item[Robustness of \GLIDE:]
Through our experiments, we provide empirical evidence substantiating the reliable nature of samples generated by \Glide~compared to \SD. 

\end{description}
\noindent

\noindent
\noindent


\section{Background and Related Works}
The diffusion model~\citep{diff_2015} is a latent variable model that can be described as a Markov chain with learned Gaussian transitions. 
It consists of two main components: the diffusion process and the reverse process. 
The reverse process is a trainable model that is trained to reduce the Gaussian noise introduced by the diffusion process systematically. 

To illustrate, consider input data represented as $x \in \R$, the approximate posterior $(q)$ is expressed by:
\begin{aequation}
q(x_t \vert x_{t-1}) := \gN(x_{t};~ \sqrt{1 - \beta_t} \cdot x_{t-1},~ \beta_tI)
\end{aequation}
which is defined as a fixed Markov chain. 
This Markov chain progressively introduces Gaussian noise to the data in accordance with a predefined schedule of variances, denoted as $\beta_1, \beta_2, \ldots, \beta_T$:
\begin{aequation}
q(x_{1:T} \vert x_{0}) := \prod^T_{t=1} ~ q(x_t \vert x_{t-1}).
\end{aequation}
Subsequently, the reverse process with trainable parameters $p_\theta(x_{0:T})$ revert the diffusion process returning the data distribution:
\begin{aequation}
p_\theta(x_{0:t}) := p(x_T) \cdot \prod ^T_{t=1} ~ p_\theta(x_{t-1} \vert x),
\end{aequation}
\begin{aequation}
\label{eq4}
    p_\theta(x_{t-1} \vert x_t) :=  \gN \left( x_{t-1};~ \mu _\theta(x_t,t),~ \Sigma_\theta(x_t,t) \right).
\end{aequation}
where $p_\theta$ contains the mean $\mu _\theta(x_t,t)$ and the variance $\Sigma_\theta(x_t,t)$, both of them are trainable models predict the value by using the current time step and the current noise. 


By fixing the forward process variances, 
Denoising Diffusion Probabilistic Models (DDPM) \citep{ho2020denoising} modify the \Eqref{eq4} to :
\begin{aequation}
\label{eq6}
    p_\theta(x_{t-1} \vert x_t) := \gN(x_{t-1};~ \mu _\theta(x_t,t),~ \sigma^2 I).
\end{aequation}
This smart design achieved higher-quality image synthesis than Generative Adversarial Networks (GANs) \citep{goodfellow2014generative}.
\newline
\newline
\noindent
\textbf{Diffusion Models and its variants:} Similar to other types of generative models ~\citep{mirza2014conditional, sohn2015learning}, the generation process can also be conditioned. 
For instance, \GLIDE~learns to generate images according to an input textual sentence on the image space, DALL·E-2 \citep{ramesh2022hierarchical} uses a DDPM to learn a prior distribution on the CLIP \citep{radford2021learning}. 
Text-to-image generation is also explored in Stable Diffusion \citep{stable_dif} and Imagen \citep{saharia2022photorealistic}. Furthermore, the release of Stable Diffusion has catalyzed a surge in diverse applications and workloads.
However, the recursive sampling process makes the diffusion model a time-consuming model. 
To address this problem, \citet{song2020denoising} proposed Denoising Diffusion Implicit Models (DDIM) \citep{song2020denoising}, a non-Markovian inference process that faster the sampling process. \citet{salimans2022progressive} propose to distill the prediction network into new networks, which progressively reduce the number of sampling steps. 
\citet{stable_dif} speed up sampling by splitting the process into a compression stage and a generation stage and applying the DDPM on the compressed (latent) space.
\newline
\newline
\noindent
\textbf{Diffusion Model lack reliable explanations:} Other than improving the application site, a sort of research notice that Diffusion Model lacks reliable explanations \citep{ning2023input,li2023alleviating,daras2023consistent}. 
This problem is called exposure bias or sampling drift. 
The researchers claim that error propagation happens to diffusion models because the models are of a cascade structure. 
\citet{li2023diffusion} further develop a theoretical framework for analyzing the error propagation of diffusion models. 
\newline
\newline
\noindent
\textbf{Impact of Seed Vector to Diffusion Models:} Recently, latent space and seed vectors have been shown to be highly correlated to the final result \citep{samuel2023norm,wu2022making, ge2023preserve}. 
Better seed vectors can consistently generate high-quality images, and even facilitate conditional control to achieve desired results \citep{mao2023guided, singh2022conditioning}.
This impact is compounded when generating the rare distribution such as rare fine-grained concepts or rare combinations \citep{liu2022compositional,zhao2019image,feng2022training,chefer2023attend}. 
\citet{samuel2023all} shows that finely selected seed vectors can generate rare distribution, which raises our interest in understanding how seed vectors behave differently.   

\section{Evaluating Reliability of Diffusion Models}

\subsection{Problem Definition}
Consider a condition-based diffusion model $G(\cdot)$ in which we generate samples $x$ using the equation $x = G(z, c)$. 
Here, $z$ is the initial seed vector following a simple and tractable normal distribution, $z \sim \gN(\mu, \alpha^2)$, as described in prior work \citep{diff_2015} and $c$ is the conditioning variable to which $x$ is closely related. 
This conditioning variable can represent various data types, such as images, sentences, or sounds. 
The strength of the correlation between $x$ and $c$ can be quantified as $p(x \vert c) \propto M(x, c)$. 
Here $M$ is a function that evaluates the variable $x$ to the conditioning variable $c$.

Further, the susceptibility of the diffusion model $G(\cdot)$ to the initial seed vector $z$ can be evaluated by manipulating $z$.
To introduce shifts to the initial seed vector $z$, we transform the initial seed vector $z$ with $\eta$, resulting in $\tilde{z} = z + \eta$.
Here, $\tilde{z}$ is the modified seed vector.
Further, the sample generated by the modified seed vector $\tilde{z}$ can be represented as $\tilde{x} = G(\tilde{z}, c)$. 
Particularly, we are interested in identifying instances where there exists a $\tilde{z}$ such that $M(\tilde{x}, c)$ is significantly lesser than $M(x, c)$.
This case implies that the modified seed vector $\tilde{z}$ breaks the diffusion model's $G(\cdot)$ relationship between generated sample $\tilde{x}$ and conditioning variable $c$, which was evident prior to the shift to initial seed vector $z$ with the generated sample $x$.


\begin{figure}[!t]
    \centering 
    \includegraphics[width=\columnwidth]{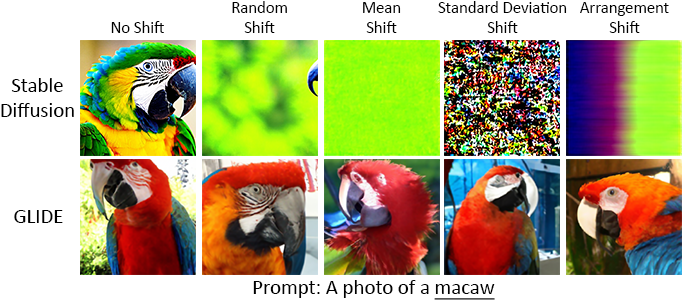}
    \caption{\textbf{Examples showcasing the impact of various shifts to initial seed vector on image generation of \SD~(Top) and \Glide(Bottom).}
    We do the following shifts to the initial seed vector (Left to Right): a) No Shift, b) \RS~$(\eta_r = 0.2)$, c) \MS~$(\eta_m = 0.2)$, d) \SDS~$(\eta_s = 0.3)$, and e) \AS~$(\eta_a = 64)$.}
    \label{fig:compare}
\end{figure}

\begin{table*}[!ht]
\centering
\caption{
\textbf{Performance of \StableDiffNew~for all shifts proposed in \Secref{shift}.}
We evaluate top-1 accuracy and top-5 accuracy calculated using pre-trained ViT and CLIP Score calculated using the OpenCLIP model.
We observe that shifts to the initial seed vector degrade the reliability of generated samples by diffusion models as the shifts increase.}
\label{tab:shift}
\resizebox{\textwidth}{!}{
    \begin{tabular}{r|l c c c c c c c c c c c c}
    \toprule
    & \multicolumn{5}{c}{\leftarrowfill (Negative Shift) } & (No Shift) & \multicolumn{5}{c}{(Positive Shift) \rightarrowfill} \\
    \midrule
    & \multicolumn{12}{c}{\textbf{Random Shift $(\eta_r)$}} \\
        & -0.30 & -0.20 & -0.15 & -0.10 & -0.05 & 0.00 & 0.05 & 0.10 & 0.15 & 0.20 & 0.30 \\
    \midrule
    Top-1 Accuracy $(\uparrow)$ & 41.8\% & 65.6\% & 69.4\% & 71.1\% & 71.2\% & 71.6\% & \x{72.9\%} & 72.1\% & 70.1\% & 65.5\% & 33.1\% \\
    Top-5 Accuracy $(\uparrow)$ & 65.1\% & 87.5\% & 89.5\% & 89.8\% & 89.8\% & 90.0\% & \x{90.7\%} & 90.6\% & 89.6\% & 86.6\% & 52.4\% \\
    CLIP Score $(\uparrow)$ & 27.7 & 31.3 & 32.2 & \x{32.6} & \x{32.6} & 32.5 & 32.4 & 32.2 & 31.8 & 31.1 & 26.7 \\
    \midrule
    & \multicolumn{12}{c}{\textbf{Mean Shift $(\eta_m)$}} \\
        &   & -0.20 & -0.15 & -0.10 & -0.05 & 0.00 & 0.05 & 0.10 & 0.15 & 0.20 &  \\
    \midrule
    Top-1 Accuracy $(\uparrow)$ &   & 12.9\% & 42.0\% & 65.4\% & 71.0\% & 71.6\% & \x{72.2\%} & 65.5\% & 33.2\% & 6.4\% &  \\
    Top-5 Accuracy $(\uparrow)$ &   & 24.6\% & 65.8\% & 87.8\% & 89.8\% & 90.0\% & \x{90.5\%} & 86.6\% & 51.8\% & 12.3\% &  \\
    CLIP Score $(\uparrow)$ &   & 22.5 & 27.6 & 31.2 & \x{32.6} & 32.5 & 32.1 & 31.1 & 26.6 & 21.7 &  \\
    \midrule
    & \multicolumn{12}{c}{\textbf{Standard Deviation Shift $(\eta_s)$}} \\
        &   &   &  -0.30 & -0.20 & -0.10 & 0.00 & 0.10 & 0.20 & 0.30 &  &  \\
    \midrule
    Top-1 Accuracy $(\uparrow)$ &  &  &  5.0\% & 41.0\% & 67.5\% & \x{71.6\%} & 62.9\% & 7.8\% & 0.0\% &  &  &  \\
    Top-5 Accuracy $(\uparrow)$ &  &  &  7.7\% & 60.7\% & 87.9\% & \x{90.0\%} & 85.6\% & 16.7\% & 0.0\% &  &  &  \\
    CLIP Score $(\uparrow)$ &  &  &  21.5 & 28.0 & 31.8 & \x{32.5} & 31.6 & 23.6 & 20.4 &  &  &  \\
    \midrule
    & \multicolumn{12}{c}{\textbf{Mixed Shift $(\eta_s, \eta_m)$}} \\
        &  &  & (-0.30, -0.15) & (-0.20, -0.10) & (-0.10, -0.05) & (0.00, 0.00) & (0.10, 0.05) & (0.20, 0.10) & (0.30,0.15) \\
    \midrule
    Top-1 Accuracy $(\uparrow)$ &  &  & 2.1\% & 23.6\% & 66.3\% & \x{71.6\%} & 65.5\% & 7.7\% & 0.0\% &  &   \\
    Top-5 Accuracy $(\uparrow)$ &  & & 3.6\% & 38.2\% & 87.6\% & \x{90.0\%} & 87.1\% & 17.1\% & 0.0\% &  &   \\
    CLIP Score $(\uparrow)$ &  & & 17.9 & 23.2 & 31.3 & \x{32.5} & 31.6 & 24.1 & 22.2 &  &   \\
    \midrule
    & \multicolumn{12}{c}{\textbf{Arrangement Shift $(\eta_a)$}} \\
        &  &  &  &  &  & 0 & 8 & 16 & 32 & 64 &  \\
    \midrule
    Top-1 Accuracy $(\uparrow)$ &  &  &  &  &  & \x{71.6\%} & 71.4\% & 56.3\% & 1.4\% & 0.0\% &  \\
    Top-5 Accuracy $(\uparrow)$ &  &  &  &  &  & \x{90.0\%} & 90.0\% & 78.3\% & 2.7\% & 0.0\% &  \\
    CLIP Score $(\uparrow)$ &  &  &  &  &  & \x{32.5} & 32.2 & 30.8 & 22.1 & 18.2 &  \\
    \bottomrule
    \end{tabular}
}
\end{table*}

\subsection{Synthetic Shifts to the Initial Seed Vector}
\label{shift}
In our investigation, we delve into the influence of the parameter $\eta$ on the generated samples through distinct transformations, comprehensively evaluating the impact of modifications to the initial seed vector. 
In order to align with the original definition from the diffusion model \citep{diff_2015}, we ensure that our modified initial vector maintains the normal distribution-like characteristics after our transformations. 
We evaluate the following five transformations: a) \RS, b) \MS, c) \SDS, d) \MixS, e) \AS~described in brief below. 
\Figref{fig:compare} shows the generated image by \StableDiff~and \GLIDE, when subjected to these transformations.

\begin{description}[style=sameline, leftmargin=*]

\item[\RS] offers a modification method that introduces randomness in the latent values. 
It incorporates randomness by sampling from a uniform distribution within the range of $0$ to $1$ and multiplying it by a scale factor $\eta_r$.
\begin{aequation}
\tilde{z}_{\text{Random}}= z + \eta_r \cdot \mathcal{U}{[0, 1]}
\end{aequation}

\item[\MS] 
represents the straightforward mean adjustment by adding a constant value $\eta_m$ to all pixels.
\begin{aequation}
\tilde{z}_{\text{Mean}}= z + \eta_m
\end{aequation}

\item[\SDS]
modifies the variance of the distribution $z$ with a scale factor $\eta_s$. 
\begin{aequation}
\tilde{z}_{\text{StandardDeviation}}= ( 1 + \eta_s) \cdot z
\end{aequation}

\item[\MixS]
combines both mean and standard deviation shifts, providing insights into their combined influence.
\begin{aequation}
\tilde{z}_{\text{Mix}}= ( 1 + \eta_s) \cdot z + \eta_m
\end{aequation}

\item[\AS]
differs from the above methods by preserving the ideal normal distribution of mean $\mu$ and standard deviation of $\alpha^2$ while locally disrupting the normality. 
It achieves this by rearranging latent values without directly altering their values. 
\begin{aequation}
\tilde{z}_{\text{Arrangement}}= T(z, \eta_a)
\end{aequation}
Here, $T$ is a sorting function that organizes the upper-left $\eta_\alpha$ submatrix. 
For instance, for \SD~the latent vector $z \in \R^{64 \times 64 \times 4}$ and when we apply \AS with $\eta_\alpha = 2$, the latent values of upper-left $2 \times 2$ submatrix along all $4$ dimensions, i.e., a total of $16$ latent values are sorted. 
\end{description}

Note that in this study, we specifically selected shifted distributions with more than $86\%$ overlap with the ideal normal distribution. 
This criterion ensures that we avoid investigating trivial cases, such as those resulting from scale issues causing the model to fail. 
Collectively, these transformations offer a comprehensive understanding of the impact of a slight shift in the initial seed vector in the resulting generated samples and the adaptivity of different diffusion models to handle such cases.

\begin{figure*}[!ht]
    \centering
    \includegraphics[width=\textwidth]{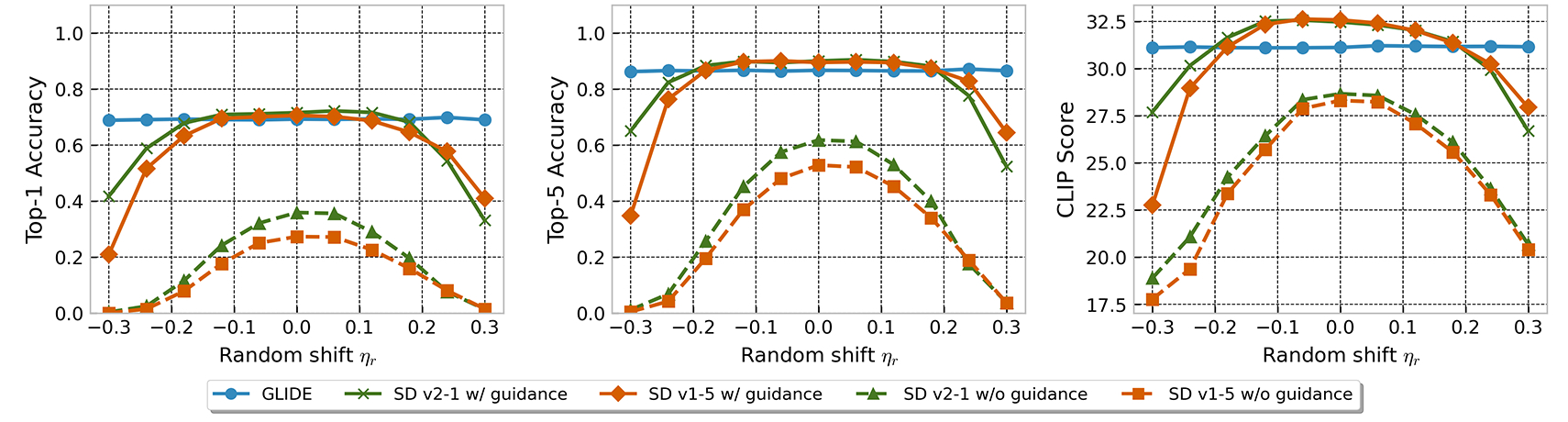}
    \caption{
    \textbf{Performance of different variants of Stable Diffusion and GLIDE for \RS.}
    We evaluate \StableDiffOld~with and without guidance, \StableDiffNew~with and without guidance and \Glide.
    Note that \Glide~shows remarkable consistency in generating samples, whereas all variants \SD~shows degradation, showing that the shifts to the initial seed vector break the correlation of the generated sample and the conditioning prompt.  
    }
    \label{fig:compare_result}
\end{figure*}

\subsection{Reliability Evaluation Framework}
To assess the reliability and correctness of generated  with shifts to the initial seed vector, we propose a reliability evaluation framework as illustrated in \Figref{fig:framework}. 
We employ a simple strategy to include the label from a dataset as a conditioning variable to the diffusion model defined as Conditioning Variable $c = $ ``A photo of a \x{\quad \quad}'' to evaluate different varieties of objects generated by diffusion models.
In this study, the \x{\quad \quad} is dynamically filled from the ImageNet-100 dataset \citep{ImageNet100}. For example, ``A photo of a \x{macaw}'', ``A photo of a \x{sea lion}'', ``A photo of a \x{crane}'' etc.
Before initiating the reverse diffusion process to generate the samples with the conditioning, the initial seed vector undergoes transformation through the shifts described in \Secref{shift}. 
Subsequently, various metrics are employed to evaluate the performance of the generated samples of diffusion models under each shift scale.



\section{Experimental Results}

\textbf{Common Experimental Setup.}
In all our experiments, we generate $100$ images for each class of ImageNet-100; thus, a total of $10,000$ images are evaluated across $100$ classes \citep{ImageNet100}. 
We compute top-1 accuracy and top-5 accuracy of the generated images using the state-of-the-art ViT-H/14 model pre-trained by SWAG \citep{singh2022revisiting} for image classification task. 
This model has an impressive top-1 accuracy of $88.55\%$ and top-5 accuracy of $98.69\%$ on the ImageNet-1k dataset. 
We also compute the CLIP score using the OpenAI-CLIP model \citep{radford2021learning}. 



\subsection{Evaluating Reliability Scores across Different Synthetic Shifts}


\textbf{Experimental setup.} 
In this experiment, we study the robustness of the \StableDiffNew~model against shifts to the initial seed vector by applying the five shifting techniques described in \Secref{shift}. 
The diffusion model setting follows the original paper \citep{stable_dif} with $50$ sampling time steps and a $7.5$ classifier-free guidance scale.
We vary the shift factor for \RS~ $\eta_r$ and \SDS~$\eta_s$ within the range $\left [ -0.3, 0.3 \right ]$.
For \MS~we vary the shift factor $\eta_m$ within the range $\left [ -0.2, 0.2 \right ]$.
In \MixS, the effects of $\eta_m$ were examined over $ \left [-0.15, 0.15 \right ]$ with $0.05$ intervals, and $\eta_s$ over $ \left [-0.3, 0.3 \right ]$ with $0.1$ intervals. 
For the \AS, we choose $\eta_a$ as $8$, $16$, $32$, and $64$.

\noindent
\textbf{Result.} 
\Tableref{tab:shift} shows the top-1 accuracy, top-5 accuracy, and CLIP Score of generated images across all types of proposed shifts and multiple scales for \StableDiffNew~model. 
Despite the pre-trained model with an $88\%$ top-1 accuracy and a $98\%$ top-5 accuracy, the situation is quite different for the \StableDiffNew~model.
It struggles to achieve only a $71\%$ top-1 accuracy and a $90\%$ top-5 accuracy. 
At the same time, we notice a drop in performance for both positive and negative shifts, and interestingly, the rate of decline is twice as fast for \MS~compared to a \RS. 
Further, we observe that introducing a positive standard deviation shift tends to deteriorate performance even more rapidly than the negative counterpart.

In both \MS~and \RS, the \StableDiffNew~demonstrates optimal performance in top-1 and top-5 accuracy (improvement of around $2\%$) with a slight positive shift ($\eta = 0.05$). 
This observation is further supported by comparing mixed shift results with standard deviation shift outcomes. 
However, there is a contrasting trend in CLIP scores where the model achieves the best CLIP score with a slight negative shift ($\eta = 0.05$).

\begin{figure*}[!ht]
    \centering
    \includegraphics[width=\textwidth]{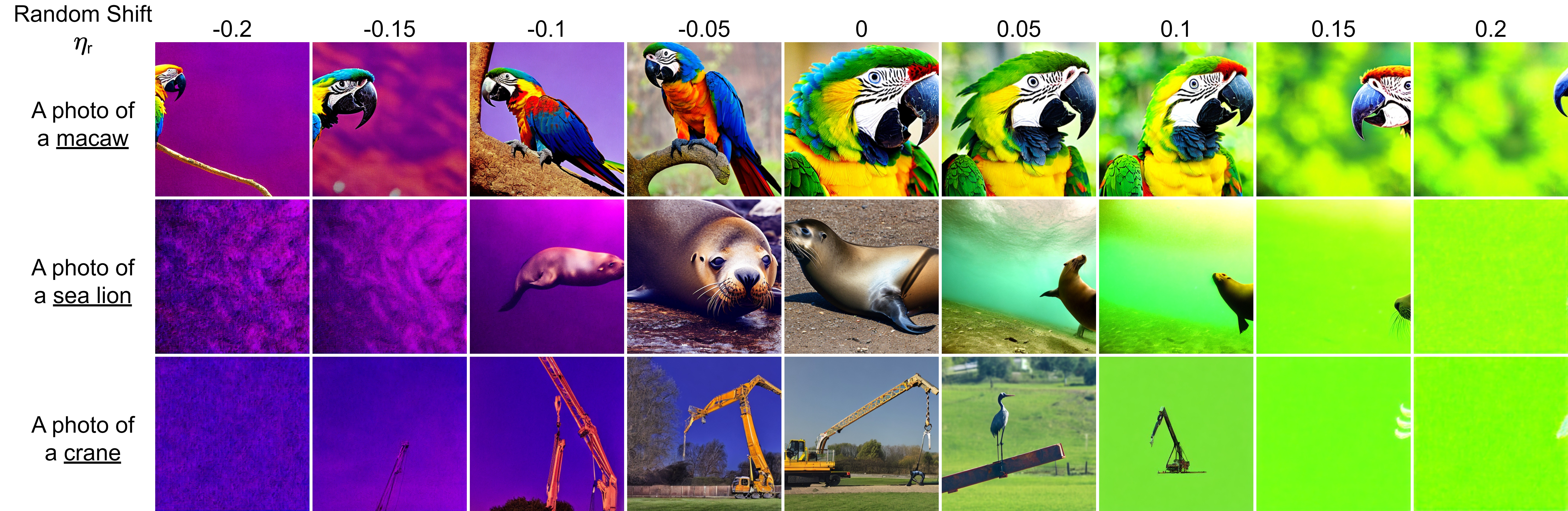}
    \caption{
    \textbf{Visual Inspection of image generated with  \RS~$(\eta_r)$ by \StableDiffNew.} 
    We give the text prompt, ``A photo of a \x{\quad}'' (row), and manipulate the initial random noise across varying levels of $\eta_r$ (column).
    Note that as $\eta_r$ deviates from zero, the images transition from accurate object representations to progressively loss of detail and color shift. 
    Correspondingly, negative shifts cause purple hues, and positive shifts result in green hues.
    }
    \label{fig:visual_sd}
\end{figure*}

\begin{figure*}[!ht]
\includegraphics[width=\textwidth]{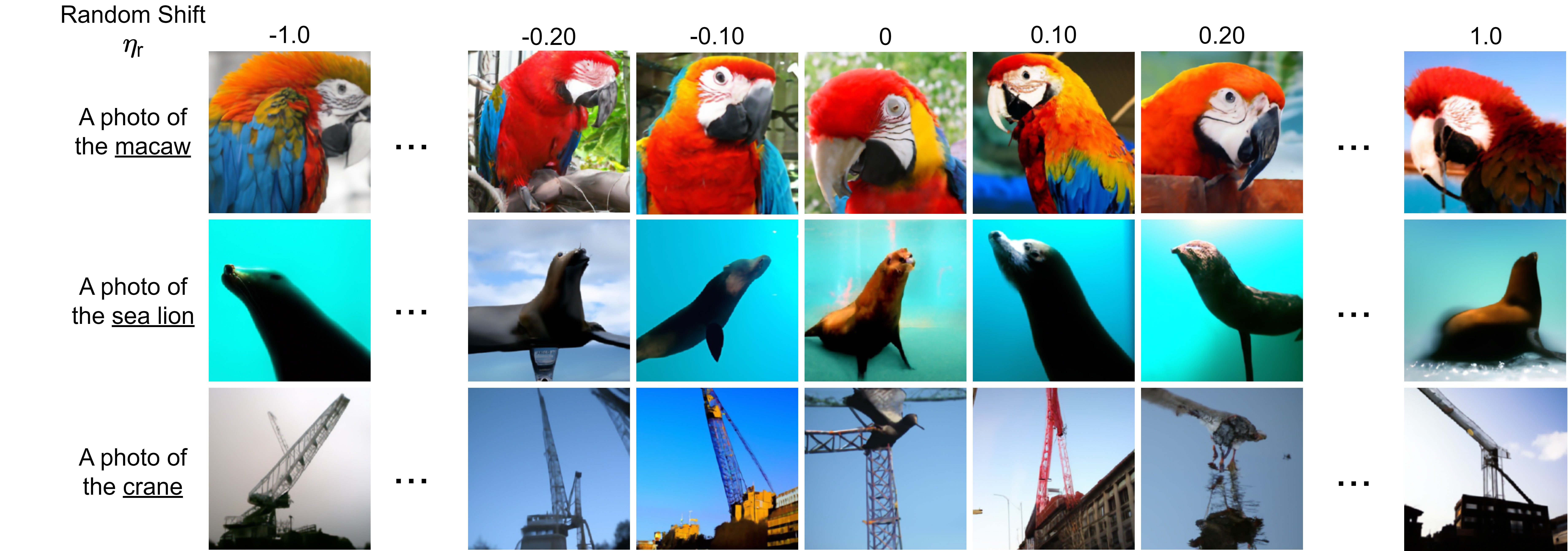}
    \caption{
    \textbf{Visual Inspection of image generated with  \RS~$(\eta_r)$ by \Glide.}
    We give the text prompt, ``A photo of a \x{\quad}'' (row), and manipulate the initial random noise across varying levels of $\eta_r$ (column).
    Unlike \SD, GLIDE exhibits consistency in generating images regardless of the degree of Random Shift applied, highlighting \Glide's superior reliability in generating images against shifts in the initial vector.}
    \label{fig:visual_glide}
\end{figure*}

\begin{figure*}[!ht]
    \centering
    \includegraphics[width=\textwidth]{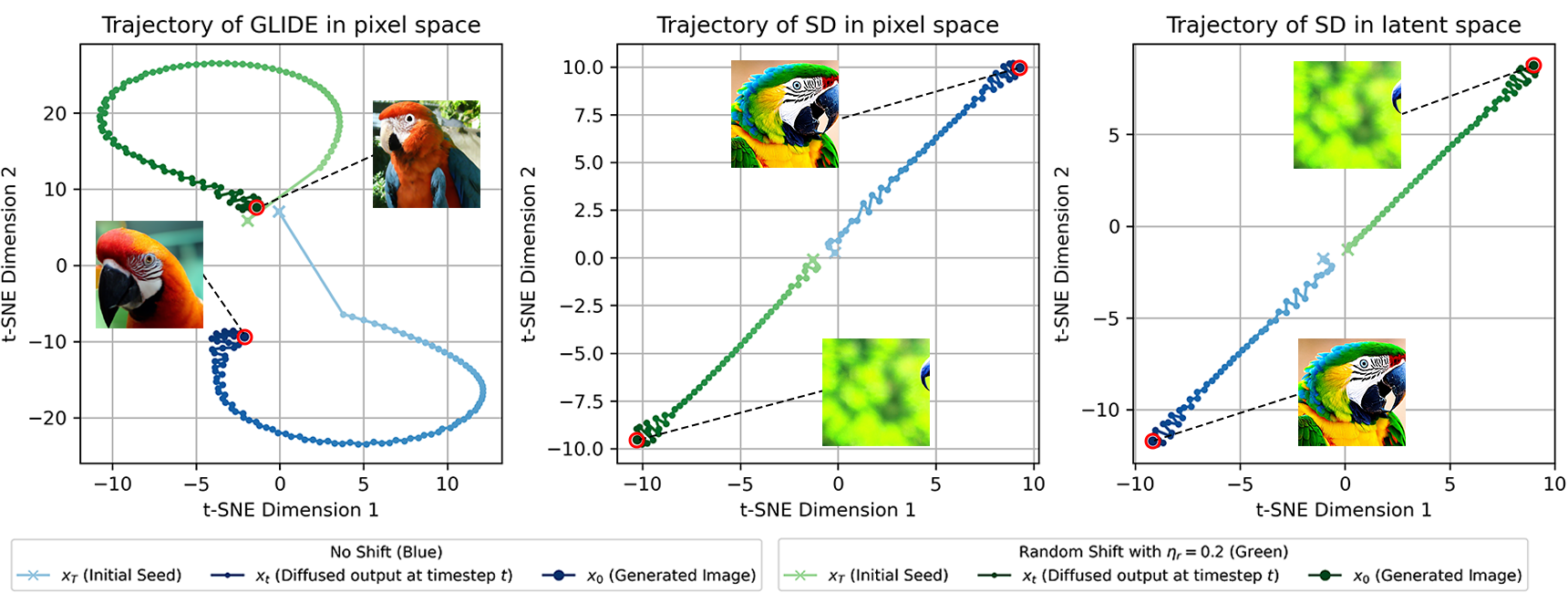}
    \caption{
    \textbf{The trajectories of the reverse diffusion process of GLIDE (Left) and Stable Diffusion (Center and Right) Models in Pixel Space (Left and Center) and Latent Space (Right) using t-SNE.}
    We plot the trajectories of the reverse diffusion process from the initial seed vector $z$ and shifted seed vector $\tilde{z}$.
    Notably, we observe a phenomenon where GLIDE's trajectory contains a large discontinuity in the early time step, 
    visible between $x_T$ (seed vector) and $x_{T-1}$ (first noise correction).
    In contrast, the Stable Diffusion does not exhibit the same behavior, and the diffusion process diverges for modified seed vector $\tilde{z}$
    compared to the initial seed vector $z$ 
    This comparison highlights the relative stability of GLIDE in generating consistent images regardless of shifts to initial seed vectors.
    }
    \label{fig:trajectory}
\end{figure*}

\subsection{Evaluating Reliability Scores for Different Diffusion Models}

\textbf{Experimental setup.}
In this experiment, we compare variants of Stable Diffusion (\StableDiffOld~with and without guidance, \StableDiffNew~with and without guidance) and GLIDE using \RS ($\eta_r$).
Although extremely rare, such a shift can occur due to sampling variability, which is possible in real-world scenarios.
We compare by varying the perturbation $\eta_r$ within the range $\left [ -0.3, 0.3 \right ]$ with an interval of $0.6$. 
For GLIDE, we follow the originally proposed setting with $150$ diffusion steps for the low-resolution image and $27$ diffusion steps for upscaling it to the high-resolution image.
For all Stable Diffusion models, we follow the original paper \citep{stable_dif} with $50$ sampling time steps and a $7.5$ classifier-free guidance scale.

\noindent
\textbf{Result.} 
\Figref{fig:compare_result} shows the performance of different diffusion models. 
The results reveal that, across all metrics evaluated, the performance of the latent-based diffusion models notably decreased as the shift increased. 
In contrast, \Glide~maintains a consistent performance by remaining unaffected by the shift irrespective of the intensity. 
Initially, with no or subtle shifts, variations of the \SD~models with guidance exhibit a slightly better performance over GLIDE. However, as the shift intensity increases, a clear divergence in performance emerges. 
The latent diffusion models demonstrate a $50\%$ drop in both top-1 and top-5 accuracy and a $30\%$ drop in CLIP Score, while \Glide's performance remains comparatively unaffected. 
For visual comparison, we provide the images generated using multiple prompts by Stable Diffusion in \Figref{fig:visual_sd} and GLIDE in \Figref{fig:visual_glide}.
We note that in the case of no shift to the initial random vector, the image quality of Stable Diffusion is better compared to GLIDE.
However, our aim is to evaluate the reliability of generated images in the context that the object given in the conditioning prompt is generated in the image rather than measuring the image quality. 

Furthermore, a comparison within the latent-based diffusion models reveals distinct behavior based on the presence of guidance. 
Models with guidance not only outperform their counterparts without guidance, which aligns with findings from previous studies \citep{DBLP:conf/icml/NicholDRSMMSC22}, but also show greater reliability to increasing shift. 
This is reflected by a slower rate of performance drop in models with guidance under subtle shifts, as opposed to the steeper decline observed in models without guidance, reaching $0\%$ in top-1 and top-5 accuracy, suggesting that the models fail to incorporate the conditioning variable in generating the outputs. 
Overall, the empirical result suggests that \Glide's reliable generations are more pronounced than that of the latent-based diffusion models.


\section{Discussion}

\begin{figure*}[!ht]
    \centering
    \includegraphics[width=\textwidth]{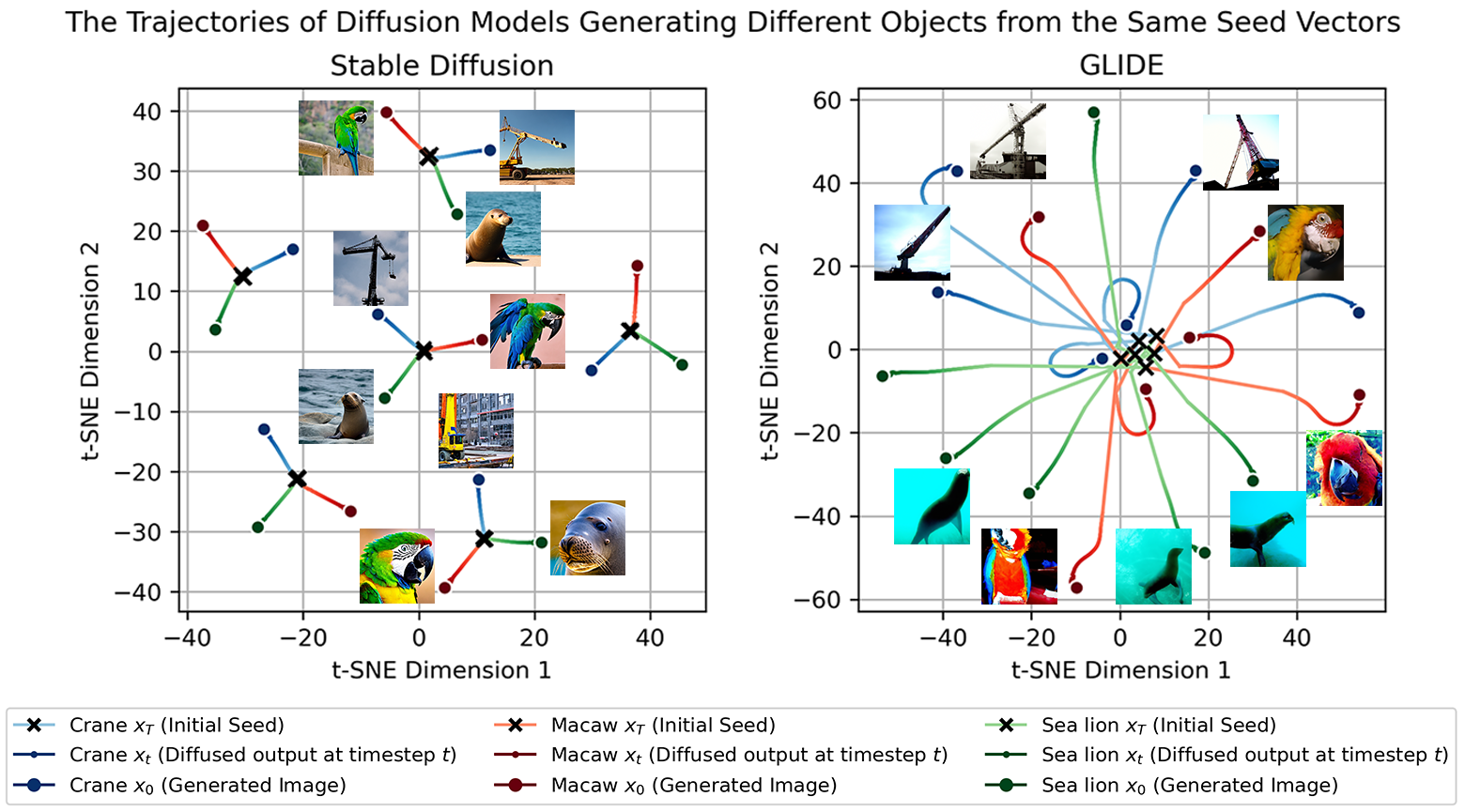}
    \caption{
    \textbf{The trajectories of reverse diffusion process of GLIDE and Stable Diffusion to generate different objects from the same initial seed vector.} The model is tasked to generate images originating from six sets of seed vectors, each set is conditioned by three prompts and each with the labels of "macaw", "crane", or "sea lion".
    We use t-SNE to reduce the visualisation of space to $2$ dimenisons.
    Originating from the same seed vectors $x_T$, Stable Diffusion's sampling trajectory is inclined to traverse toward a local desired output. 
    GLIDE, by comparison, tends to diverge, with its trajectory extending relatively farther from the seed vector toward the images $x_0$.
    }
    \label{fig:firework}
\end{figure*}

\subsection{Extended Analysis of \SD~Model}
\textbf{Disparities in Class-wise Reliable Generations:} 
The visualization in the \Figref{fig:visual_sd} distinctly showcases the varying levels of class-wise reliable generations exhibited by the model.
Notably, even when the initial random vector is subjected to a \RS~with $\eta_r = 0.15$, the model adeptly generates an image of a ``\x{macaw}'' with remarkable precision, isolating the subject from its background. 
In stark contrast, attempts to generate an image of a ``\x{crane}'' under similar conditions result in the object not being generated by the model. 
This discrepancy aligns with the observations of \citet{samuel2023all}, who underscore the notion that common concepts are reliably generated across a wider range of initial seed vectors. 
In contrast, generating images representing rare concepts demands a meticulous selection of initial seed vectors, and a slight disturbance may break the diffusion model to generate the said object in the conditioning variable.

\textbf{Unintended Positional Shifts:} 
The complexity of the impact of shifts on the initial seed vector is further revealed when we notice the generated images of ``\x{macaw}'' in \Figref{fig:visual_sd}. 
It is intriguing to note that the position of the generated object is strongly tied to the direction of the \RS. 
Positive \RS~ gradually displaces the ``\x{macaw}'' to the right, eventually causing it to vanish from the frame on the right side. 
Conversely, negative shifts induce a slow leftward movement of the ``\x{macaw}''. 
This observation invites further exploration into the realm of position control \citep{mao2023guided}, suggesting that manipulation of the initial seed could yield nuanced control over the final position of generated objects.

\textbf{Slight \RS~ boost performance:} 
Surprisingly, as evident in \Tableref{tab:shift}, \SD~outperforms its counterparts in cases arising from the slight shift from the standard normal distribution. 
In the context of \RS~setting, \SD~attains the best top-1 and top-5 accuracy, particularly excelling in the presence of a slight positive shift $\eta_r = \eta_m = 0.05$.
This noteworthy trend is also observed in \MS~experiments.
Furthermore, when comparing \SDS~and \MixS~scenarios, employing a mere \SDS of $\eta_s = 0.1$ yields a top-1 Accuracy of $62.0\%$ and a top-5 accuracy of $86.6\%$. 
Intriguingly, introducing a slight positive shift of $\eta_m = 0.05$, creating a \MixS, significantly enhances performance to $65.5\%$ top-1 and $87.1\%$ top-5 accuracy. 
This observation suggests that shifts, when strategically applied, can propel the initial seed vector towards a more favorable starting position, thereby boosting overall performance as noted by \citep{samuel2023all,samuel2023norm,mao2023guided}.

\subsection{Robustness of \Glide~to Synthetic Shifts}
To discuss the GLIDE's highly reliable generations compared to the Stable Diffusion model, we display the difference in sampling trajectories of the reverse diffusion process for both models with Figures \ref{fig:trajectory} and \ref{fig:firework}.



\Figref{fig:trajectory} shows the trajectories of the reverse diffusion process of GLIDE and Stable Diffusion using t-SNE. We plot the difference in trajectories of the reverse diffusion process for initial seed vector $z$ and shifted seed vector $\tilde{z}$.
Notably, GLIDE's sampling trajectory exhibits a significant discontinuity in the early time step, a phenomenon we coin as Early Steps Discontinuity.
Such discontinuity hinted that GLIDE "pulls" the shifted seed vectors back towards a standard normal distribution, maintaining its output quality. 
Consequently, this could account for GLIDE's relative stability in consistently generating images despite transformations to initial shift vectors.
On the contrary, the Stable Diffusion model does not display the same behavior.
Its sampling trajectory with the shifted seed $\tilde{z}$ diverges when compared to the trajectory of $z$.



We further discuss the contrasting trajectories of the reverse diffusion process of GLIDE and Stable Diffusion for generating different objects using conditioning variable $c$ in response to the same seed vectors with \Figref{fig:firework}.
Originating from the same seed vectors, Stable Diffusion’s sampling trajectory tends to traverse towards a local desired output, indicating that it is more inclined to search locally from the random seed vector for the desired output. 
On the other hand, GLIDE sampling process exhibits a divergent pattern, with its trajectory extending relatively further from the seed vector toward the generated images, reflecting a broader exploration of the solution space.

We speculate that these factors contribute to the generation of non-reliable images by \SD~when initial seed vector $z$ is transformed using $\eta$ compared to the generation of reliable images by \Glide.
In summary, this analysis sheds light on the differences in \SD~and \Glide~ the reasoning for the more reliable generation of images by \Glide~compared to \SD, emphasizing the importance of the chosen diffusion process and training strategy in influencing the overall performance and resilience of diffusion models.

\section{Conclusion}

This paper conducts a comprehensive analysis of the generated images by the Stable Diffusion and GLIDE models when the initial seed vector of diffusion models is subjected to transformations.
Our findings indicate that the state-of-the-art latent-based diffusion model \SD~struggles to effectively manage diverse shifts generating objects irrespective of the initial seed vector as the scale of the shift increases. 
On the other hand, a relatively older diffusion model \Glide~demonstrates a higher resiliency to handle such shifts to the initial seed vector. 
Through a combination of experimental and theoretical approaches, we identify and elucidate the factors contributing to \Glide's superior reliable generations compared to \SD. 
We anticipate that our work will serve as a foundational resource for researchers aiming to design diffusion models that are simultaneously stable and reliable in generating objects.

{
    \small
    \bibliographystyle{iclr2024_conference}
    \bibliography{_arxiv}
}

\clearpage
\setcounter{page}{1}
\setcounter{section}{0}
\renewcommand{\thesection}{\Alph{section}}

\section{Analysis of Overlap of Shifted Distributions with respect to Standard Normal Distribution}

\citet{diff_2015} assumes that the initial seed vector for diffusion models be sampled from the standard normal distribution for generating high-fidelity images from diffusion models.
We propose to use the synthetic shifts, which follow a normal distribution but keep the difference to the standard normal distribution as small as possible. 
Moreover, we do not want to shift the transformations too drastically because it goes beyond the range of learned latent values by diffusion models. 
When designing the experiments, we carefully considered the distribution overlap to ensure that the transformations are not too drastic and have a high overlap with standard normal distribution. 
The overlap percentage of shifted distribution to standard normal distribution suggested by \citep{diff_2015} can be referred in \Tableref{tab:overlap}.
Most cases of scaled shifts have over $90\%$ overlap with the standard normal distribution, while all the evaluated scaled shifts have an overlap of more than $80\%$ with the standard normal distribution. 
This implies that the latent values in the shifted seed vectors have a relatively high likelihood of being sampled from the ideal normal distribution.

\begin{table*}[!b]
\centering
\caption{\textbf{Distribution overlap between initial and shifted seed vectors.} The table shows how the shifted seed vectors differ from the initial seed vectors. 
Random Shift and Mean Shift are more than $90\%$ overlap with the initial seed vector. Standard Deviation Shift and Mixed Shift are all higher than $80\%$. 
Regarding Arrangement Shift, since we are only switching the values, it has $100\%$ overlap with the initial seed vector, regardless of $\eta_a$. }
\label{tab:overlap}
\resizebox{\textwidth}{!}{
    \begin{tabular}{r|l c c c c c c c c c c c c}
    \toprule
    & \multicolumn{5}{c}{\leftarrowfill (Negative Shift) } & (No Shift) & \multicolumn{5}{c}{(Positive Shift) \rightarrowfill} \\
    \midrule
    & \multicolumn{12}{c}{\textbf{Random Shift $(\eta_r)$}} \\
        & -0.30 & -0.20 & -0.15 & -0.10 & -0.05 & 0.00 & 0.05 & 0.10 & 0.15 & 0.20 & 0.30 \\
    \midrule
    Overlap of Distribution $(\uparrow)$ & 94.02\% & 96.01\% & 97.01\% & 98.00\% & 99.00\% & 100.0\% & 99.00\% & 98.00\% & 97.01\% & 96.01\% & 94.02\% \\
    \midrule
    & \multicolumn{12}{c}{\textbf{Mean Shift $(\eta_m)$}} \\
        &   & -0.20 & -0.15 & -0.10 & -0.05 & 0.00 & 0.05 & 0.10 & 0.15 & 0.20 &  \\
    \midrule
    Overlap of Distribution $(\uparrow)$ &   & 92.03\% & 94.02\% & 96.01\% & 98.00\% & 100.0\% & 98.00\% & 96.01\% & 94.02\% & 92.03\% &  &  \\
    \midrule
    & \multicolumn{12}{c}{\textbf{Standard Deviation Shift $(\eta_s)$}} \\
        &   &   &  -0.30 & -0.20 & -0.10 & 0.00 & 0.10 & 0.20 & 0.30 &  &  \\
    \midrule
    Overlap of Distribution $(\uparrow)$ &  &  & 83.00\% & 89.24\% & 94.90\% & 100.0\% &  95.39\% & 91.20\% & 87.37\% &  &  &&  \\
    \midrule
    & \multicolumn{12}{c}{\textbf{Mixed Shift $(\eta_s, \eta_m)$}} \\
        &  &  & (-0.30, -0.15) & (-0.20, -0.10) & (-0.10, -0.05) & (0.00, 0.00) & (0.10, 0.05) & (0.20, 0.10) & (0.30,0.15) \\
    \midrule
    Overlap of Distribution $(\uparrow)$ &  &  & 82.00\% & 88.58\% & 94.60\% & 100.0\% & 95.10\% & 90.66\% & 86.60\% & &  &   \\
    \midrule
    & \multicolumn{12}{c}{\textbf{Arrangement Shift $(\eta_a)$}} \\
        &  &  &  &  &  & 0 & 8 & 16 & 32 & 64 &  \\
    \midrule
    Overlap of Distribution $(\uparrow)$ &  &  &  &  &  & 100.0\% & 100.0\% & 100.0\% & 100.0\% & 100.00\% &  \\
    \bottomrule
    \end{tabular}
}
\end{table*}

\section{Extended Analysis for Robustness of \Glide~to Synthetic Shifts}

\begin{figure*}[!ht]
    \centering
    \includegraphics[width=0.9\textwidth]{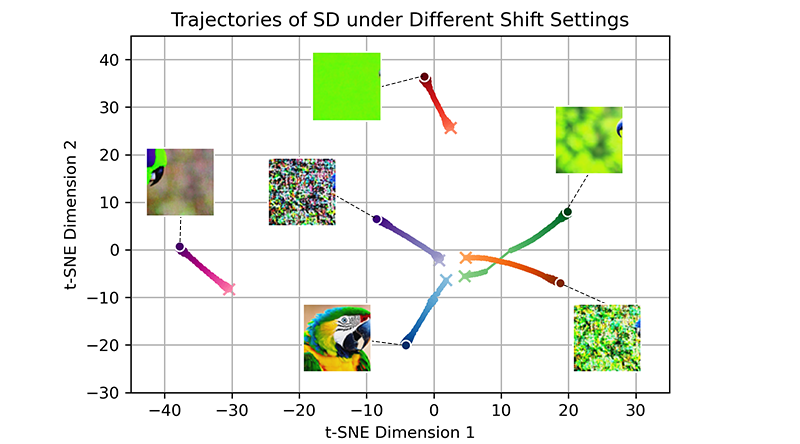} \\
    \includegraphics[width=0.9\textwidth]{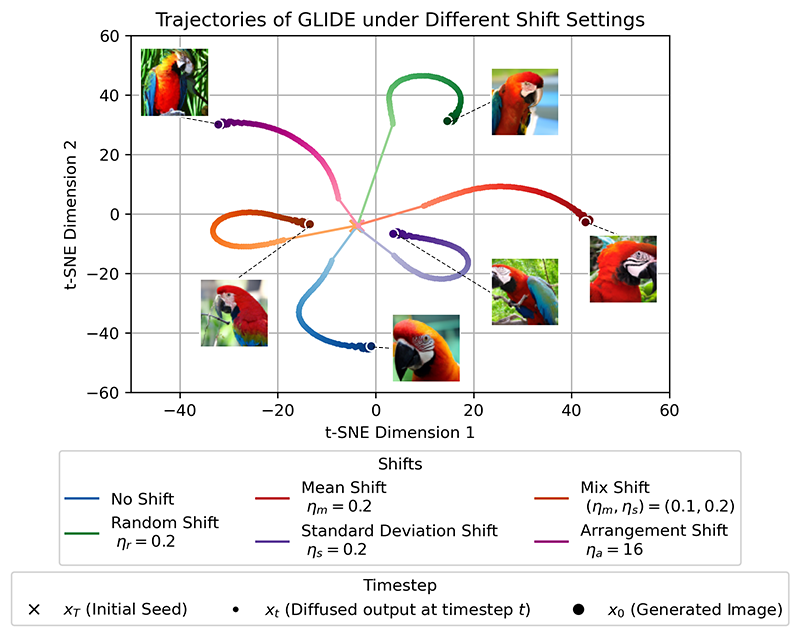}
    \caption{
    \textbf{Visual Inspection of reverse diffusion process trajectories of Stable Diffusion and GLIDE with shifts.} 
    }
    \label{fig:appendix_trajectory}
\end{figure*}

\begin{figure*}[!ht]
    \centering
    Reverse Diffusion Process of \SD\\
    \includegraphics[width=0.55\textwidth]{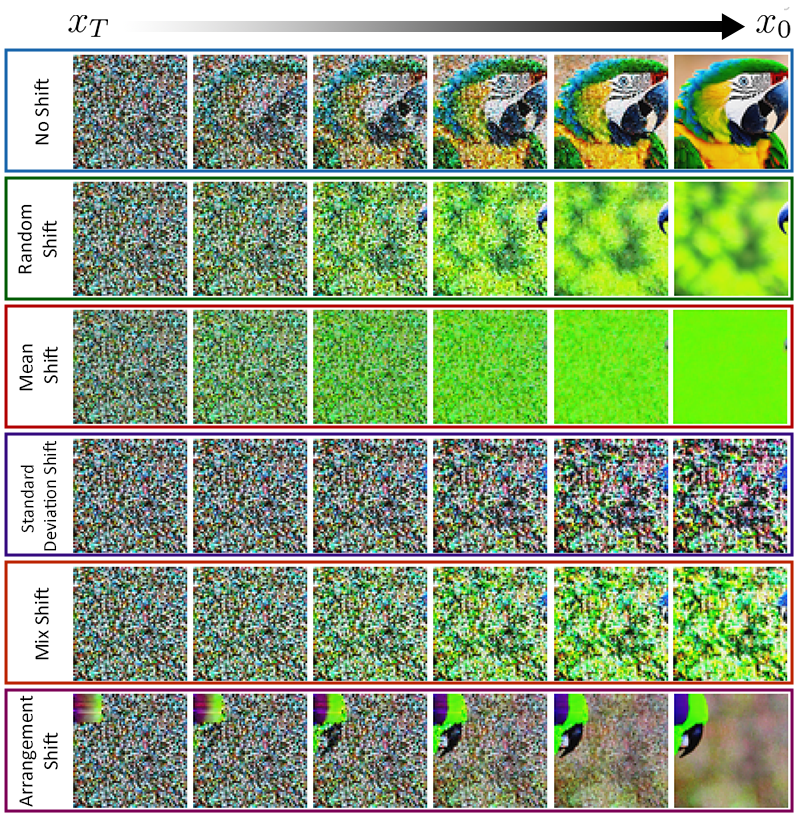}
    \vspace{0.1cm}
    \hrule
    \vspace{0.1cm}
    Reverse Diffusion Process of \Glide\\
    \includegraphics[width=0.55\textwidth]{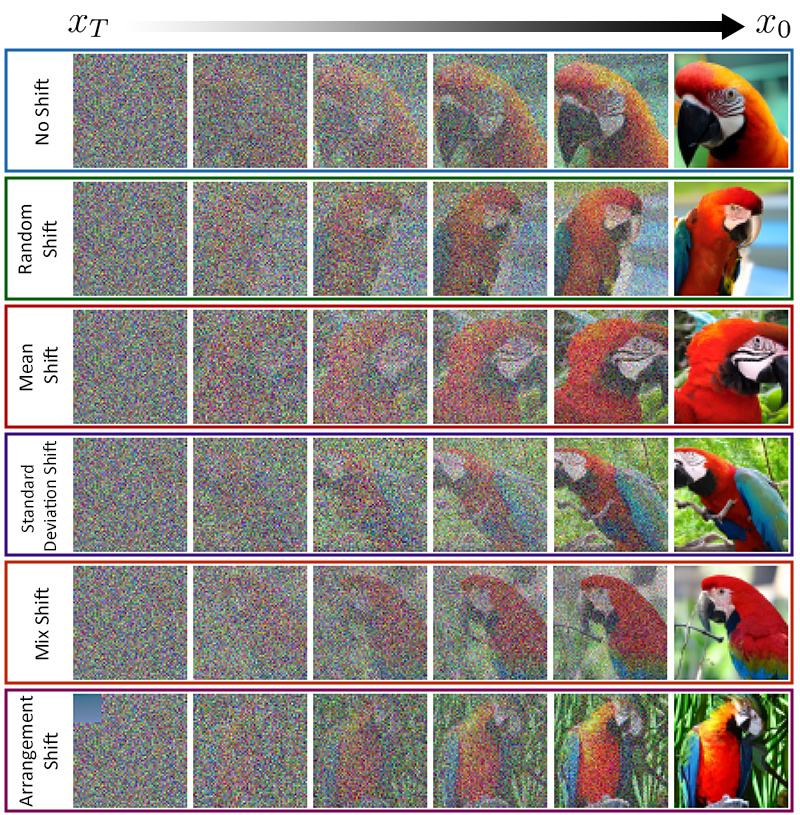}
    \caption{\textbf{Visual Inspection of the images generated in the reverse diffusion process of Stable Diffusion and GLIDE with shifts.} }
    \label{fig:appendix_sampling}
\end{figure*}

Commencing with an initial seed vector $z$ (depicted in the ``blue'' cross), distinct shifts are sequentially applied (depicted in various colored crosses). 
The initial seed vector $z$, and the shifted seed vector $\tilde{z}$, follow distinct reverse diffusion process trajectories, as illustrated in \Figref{fig:appendix_trajectory}. 
The temporal evolution of the reverse diffusion process is displayed in \Figref{fig:appendix_sampling}.

In \Figref{fig:appendix_trajectory}, despite the visual similarity among trajectories, they yield entirely different outputs. 
Each time step closely approximates the preceding one, and the final output remains in proximity to the initial points, a phenomenon corroborated by \Figref{fig:appendix_sampling} \figtop.
This figure underscores that, early on, each path outlines the intended content to be generated. 
Subsequently, additional details are progressively incorporated to enhance image clarity. 
However, it is crucial to note that even a seemingly negligible perturbation can mislead Stable Diffusion, leading to the generation of unintended content. 

In the \Figref{fig:appendix_sampling} \figbottom, it becomes evident that a substantial discontinuity emerges in the early time steps regardless of the starting point. 
This discontinuity is crucial in steering the shifted seed vector to a more favorable starting point. 
This phenomenon is particularly pronounced in the \AS~case. 
Examining the last row of \Figref{fig:appendix_sampling}\figbottom, we observe an arrangement patch situated at the top left during the initial time step. 
However, after a few subsequent time steps, it vanishes. 
This observation underscores the significance of discontinuity in effectively managing shifted seed vectors.

\section{Visual Examples of Image Generation by Different Diffusion Models for Synthetic Shifts}

\begin{figure*}[!ht]
    \centering
    \includegraphics[width=0.94\textwidth]{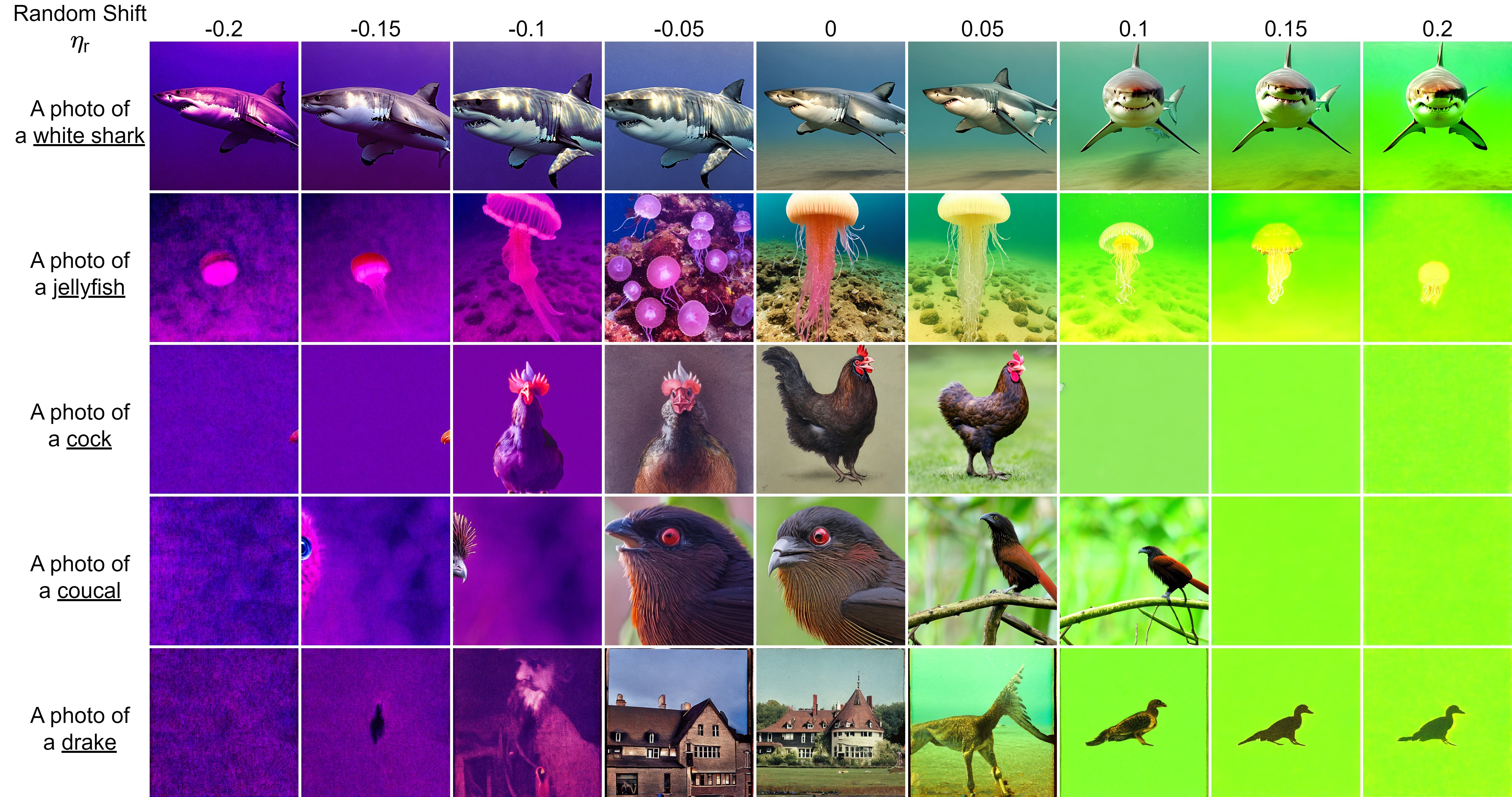}
    \caption{\textbf{Visual Inspection of image generated with \RS~$(\eta_r)$ by \StableDiffNew.}
    }
    \label{fig:appendix_sd_rs}
\end{figure*}

\begin{figure*}[!ht]
    \centering
    \includegraphics[width=0.94\textwidth]{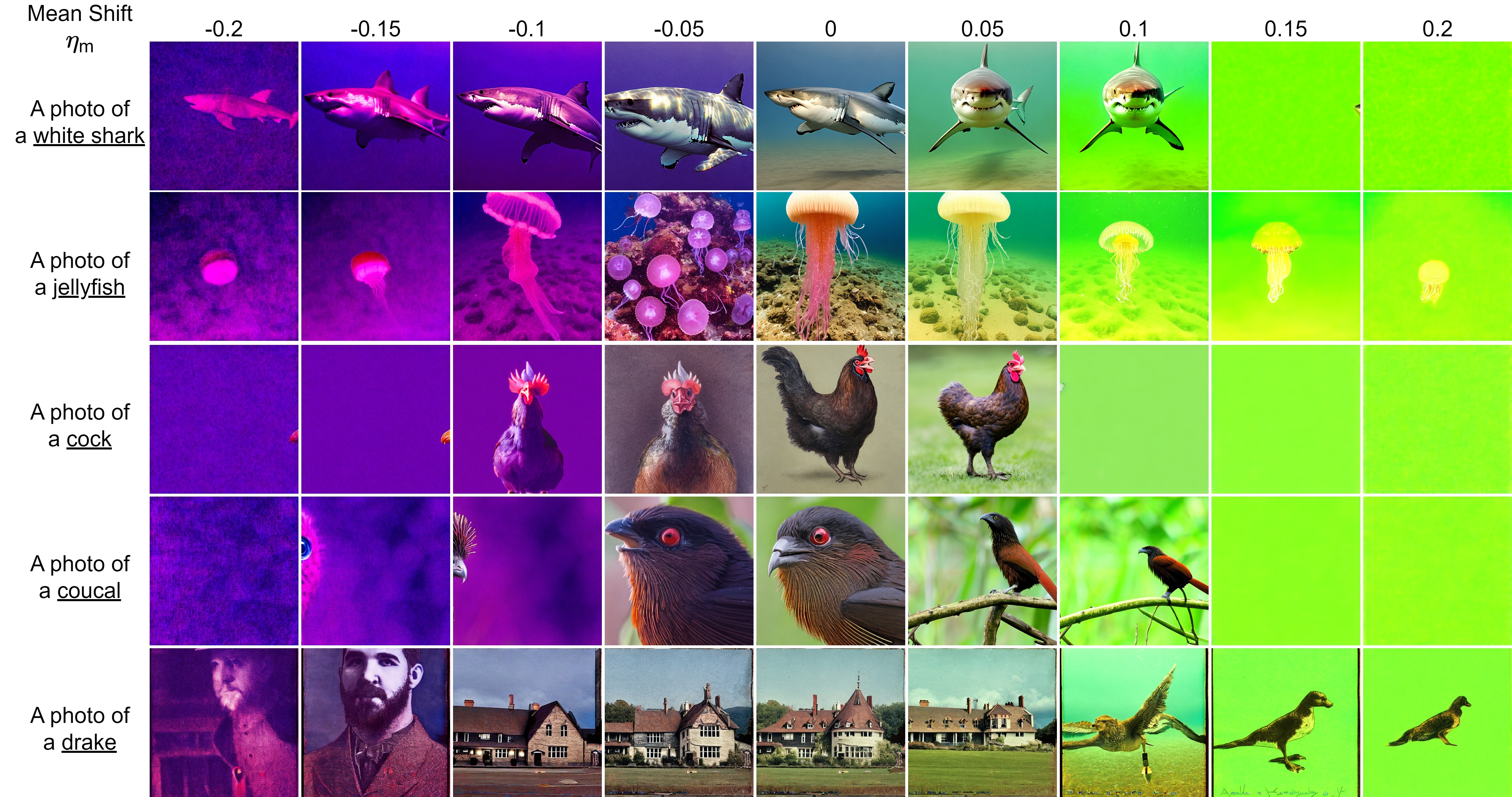}
    \caption{\textbf{Visual Inspection of image generated with \MS~$(\eta_m)$ by \StableDiffNew.} }
    \label{fig:appendix_sd_ms}
\end{figure*}

\begin{figure*}[!ht]
    \centering
    \includegraphics[width=0.8\textwidth]{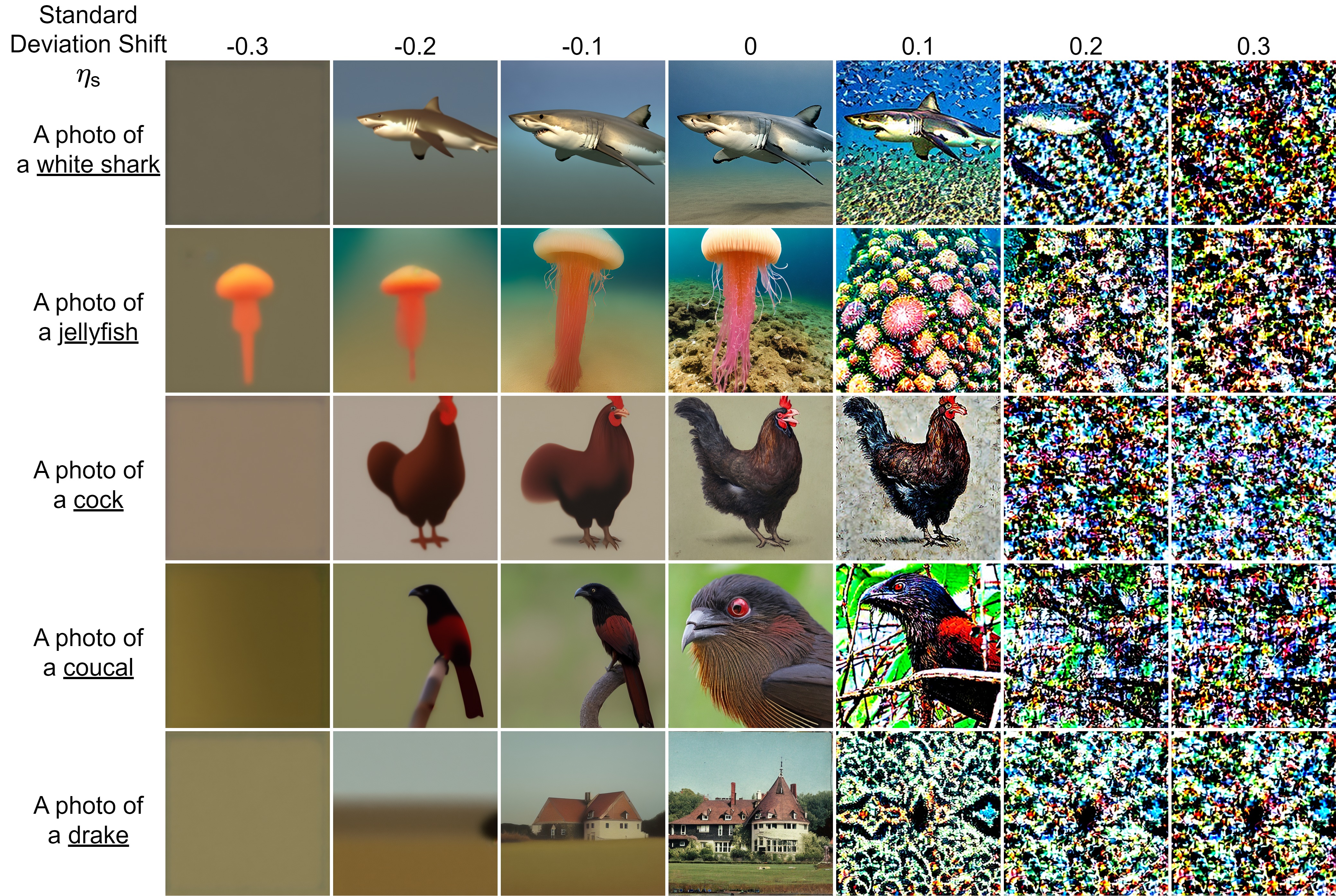}
    \caption{\textbf{Visual Inspection of image generated with \SDS~$(\eta_s)$ by \StableDiffNew.}}
    \label{fig:appendix_sd_sds}
\end{figure*}

\begin{figure*}[!ht]
    \centering
    \includegraphics[width=0.75\textwidth]{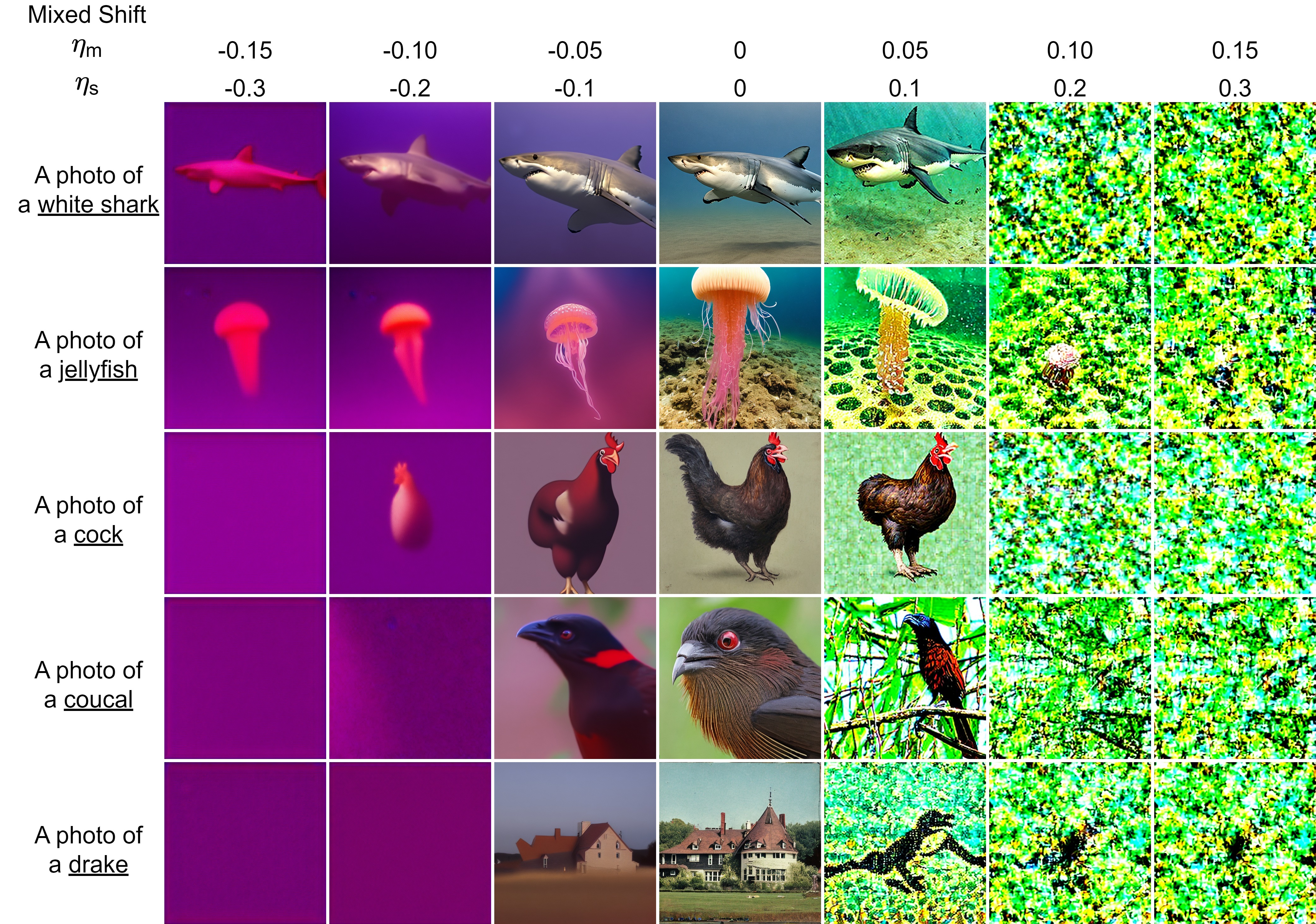}
    \caption{\textbf{Visual Inspection of image generated with \MixS~ by \StableDiffNew.}}
    \label{fig:appendix_sd_mixs}
\end{figure*}

\begin{figure*}[!ht]
    \centering
    \includegraphics[width=0.63\textwidth]{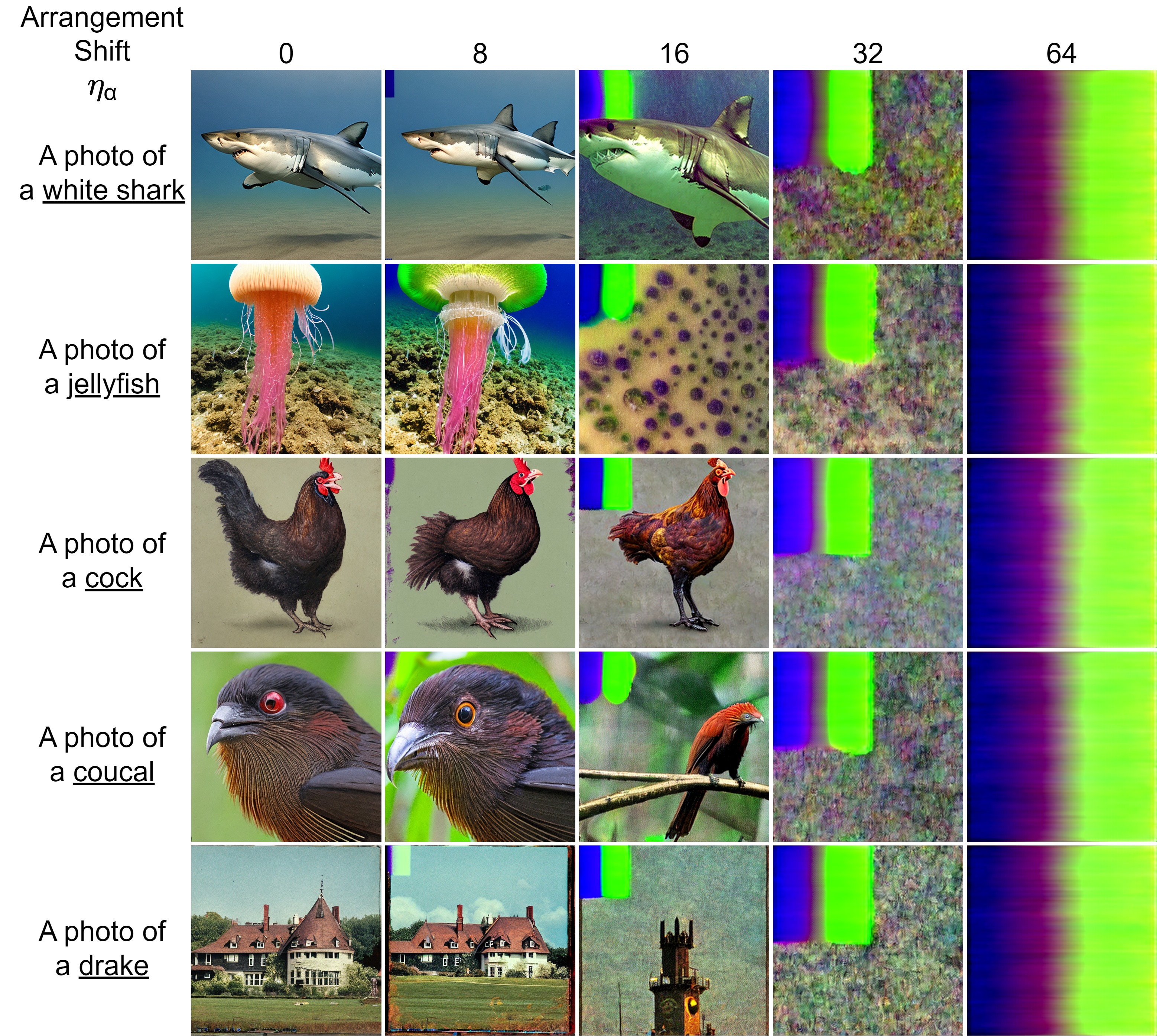}
    \caption{\textbf{Visual Inspection of image generated with \AS~ by \StableDiffNew.}}
    \label{fig:appendix_sd_as}
\end{figure*}

The outcomes of both Stable Diffusion and GLIDE shifts are illustrated in the following \Figsref{fig:appendix_sd_rs}{fig:appendix_glide_as}. 
Notably, GLIDE consistently performs uniformly across various shifts. 
In both Random Shift (\Twofigref{fig:appendix_sd_rs}{fig:appendix_glide_rs}) and Mean Shift (\Twofigref{fig:appendix_sd_ms}{fig:appendix_glide_ms}) scenarios, Stable Diffusion demonstrates similar performance. 
As $\eta_r$ or $\eta_m$ deviates from zero, there is a perceptible shift in the images from accurate depictions of objects to a gradual loss of detail and a color shift.
Negative shifts result in the presence of purple tones, while positive shifts bring about green tones.
Regarding Standard Deviation Shift (\Twofigref{fig:appendix_sd_sds}{fig:appendix_glide_sds}), positive shifts introduce noise and challenge item identification.
Conversely, negative shifts reduce detail, as exemplified in the ``A photo of a \x{cock}'' row, where feather texture and cock face detail diminish with further negative shifts. 
Mixed Shift (\Twofigref{fig:appendix_sd_mixs}{fig:appendix_glide_mixs}) combines both effects, altering color and details. 
Regarding Arrangement Shift (\Twofigref{fig:appendix_sd_as}{fig:appendix_glide_as}), arranging $8$ pixels from the top has no adverse impact on image quality. However, increasing it to $16$ pixels leads to the inability to generate the left top position and further increases blur in the remaining portion. 
Starting from arranging $32$ pixels, no image can be generated. 
These systematic experiments offer insights into how Stable Diffusion responds to perturbations across various scenarios.

\clearpage
\begin{figure*}[!ht]
    \centering
    \includegraphics[width=0.94\textwidth]{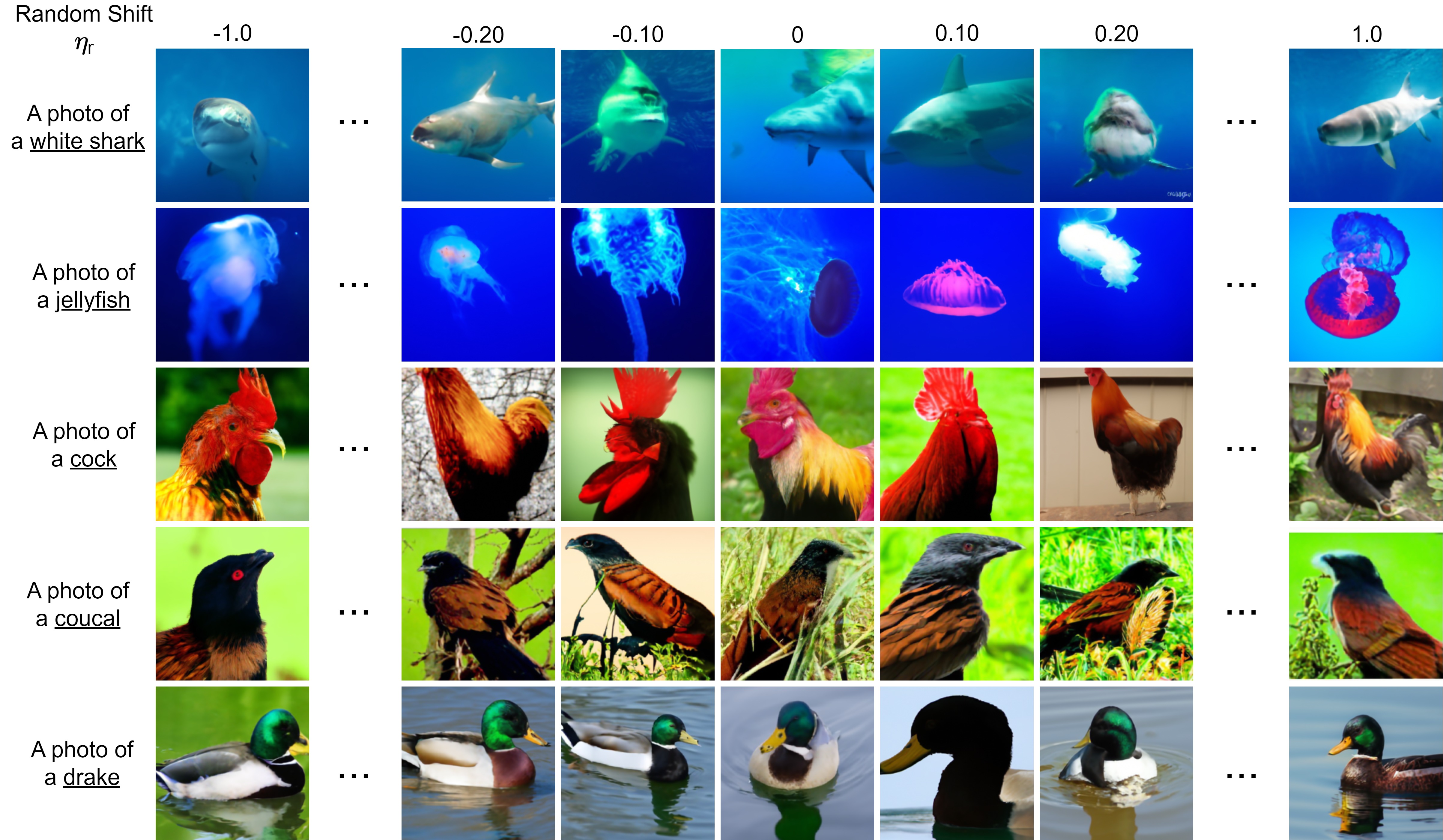}
    \caption{
    \textbf{Visual Inspection of image generated with \RS~$(\eta_r)$ by \Glide.} 
    }
    \label{fig:appendix_glide_rs}
\end{figure*}

\clearpage
\begin{figure*}[!ht]
    \centering
    \includegraphics[width=0.94\textwidth]{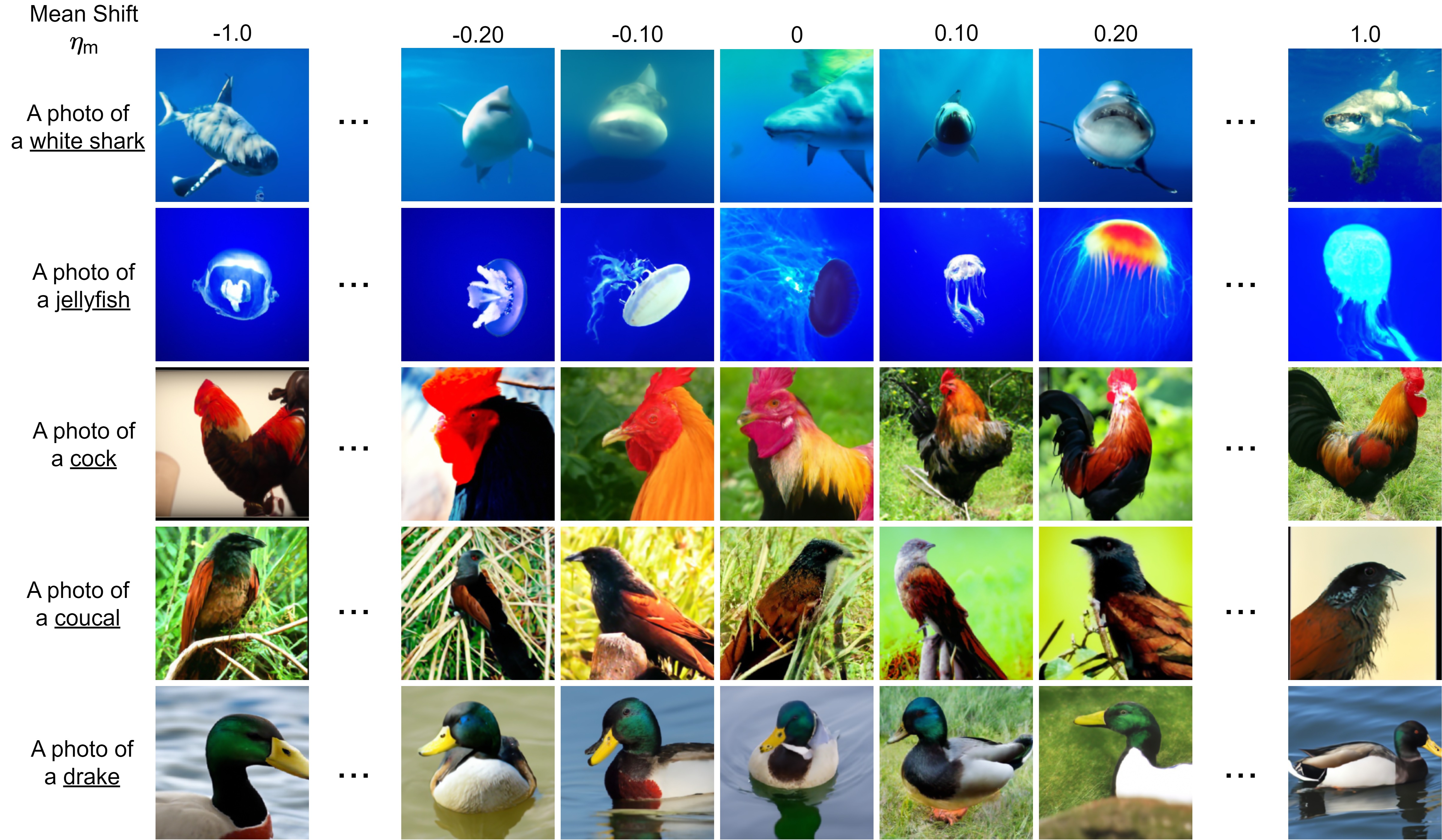}
    \caption{
    \textbf{Visual Inspection of image generated with \MS~$(\eta_m)$ by \Glide.}}
    \label{fig:appendix_glide_ms}
\end{figure*}

\clearpage
\begin{figure*}[!ht]
    \centering
    \includegraphics[width=0.94\textwidth]{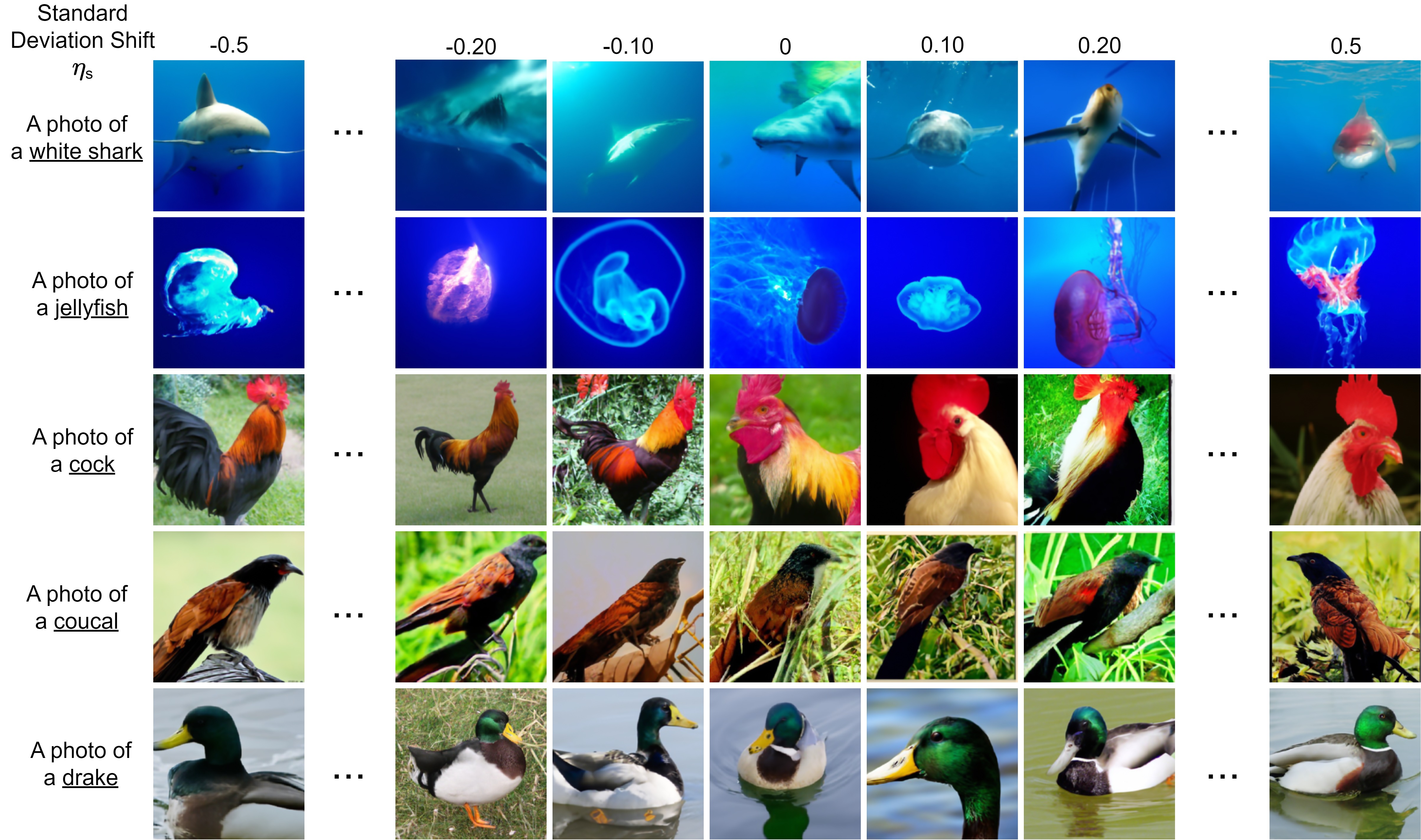}
    \caption{
    \textbf{Visual Inspection of image generated with \SDS~$(\eta_s)$ by \Glide.}}
    \label{fig:appendix_glide_sds}
\end{figure*}

\clearpage
\begin{figure*}[!ht]
    \centering
    \includegraphics[width=0.94\textwidth]{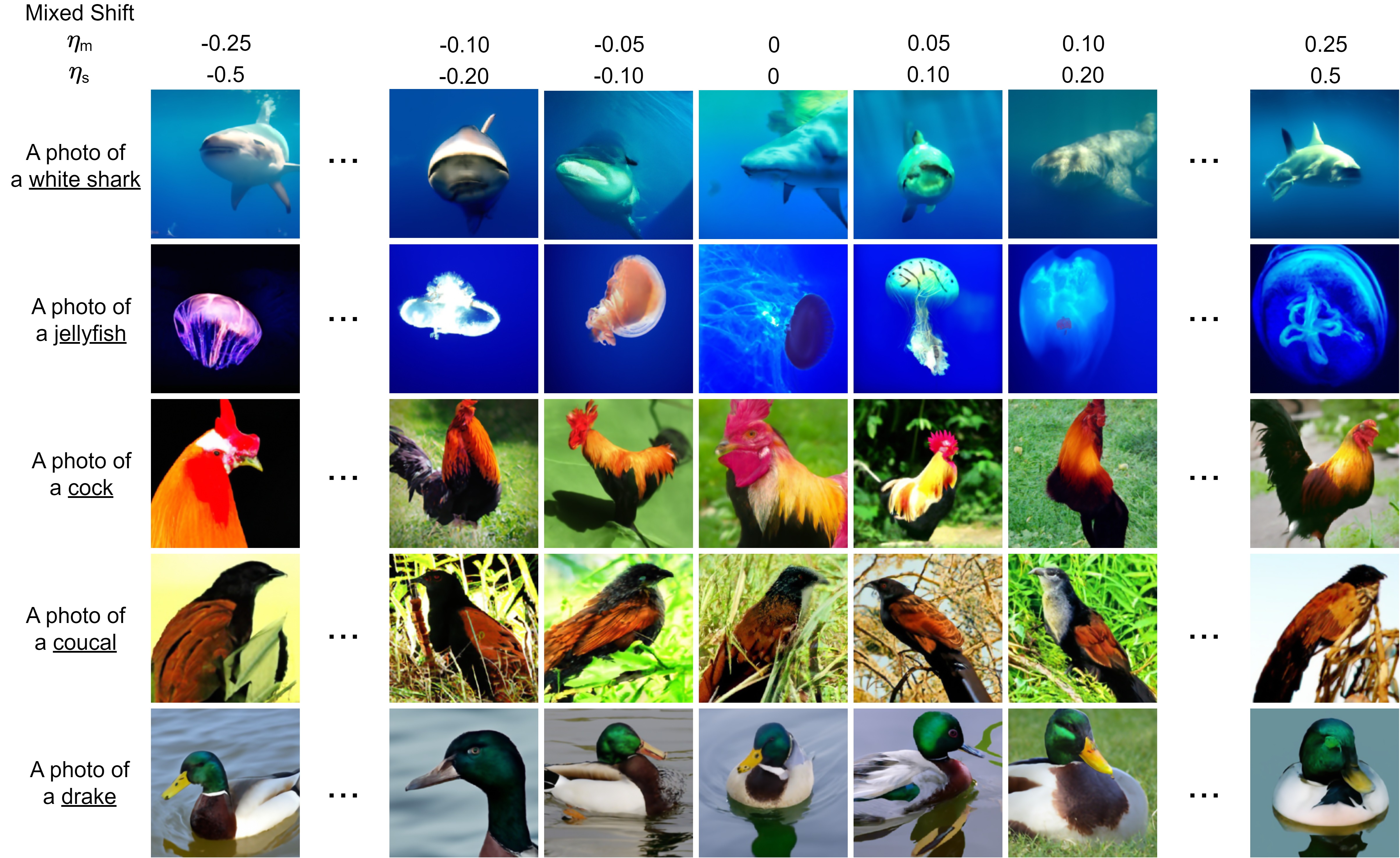}
    \caption{
    \textbf{Visual Inspection of image generated with \MixS~ by \Glide.}}
    \label{fig:appendix_glide_mixs}
\end{figure*}
\clearpage
\begin{figure*}[!ht]
    \centering
    \includegraphics[width=0.63\textwidth]{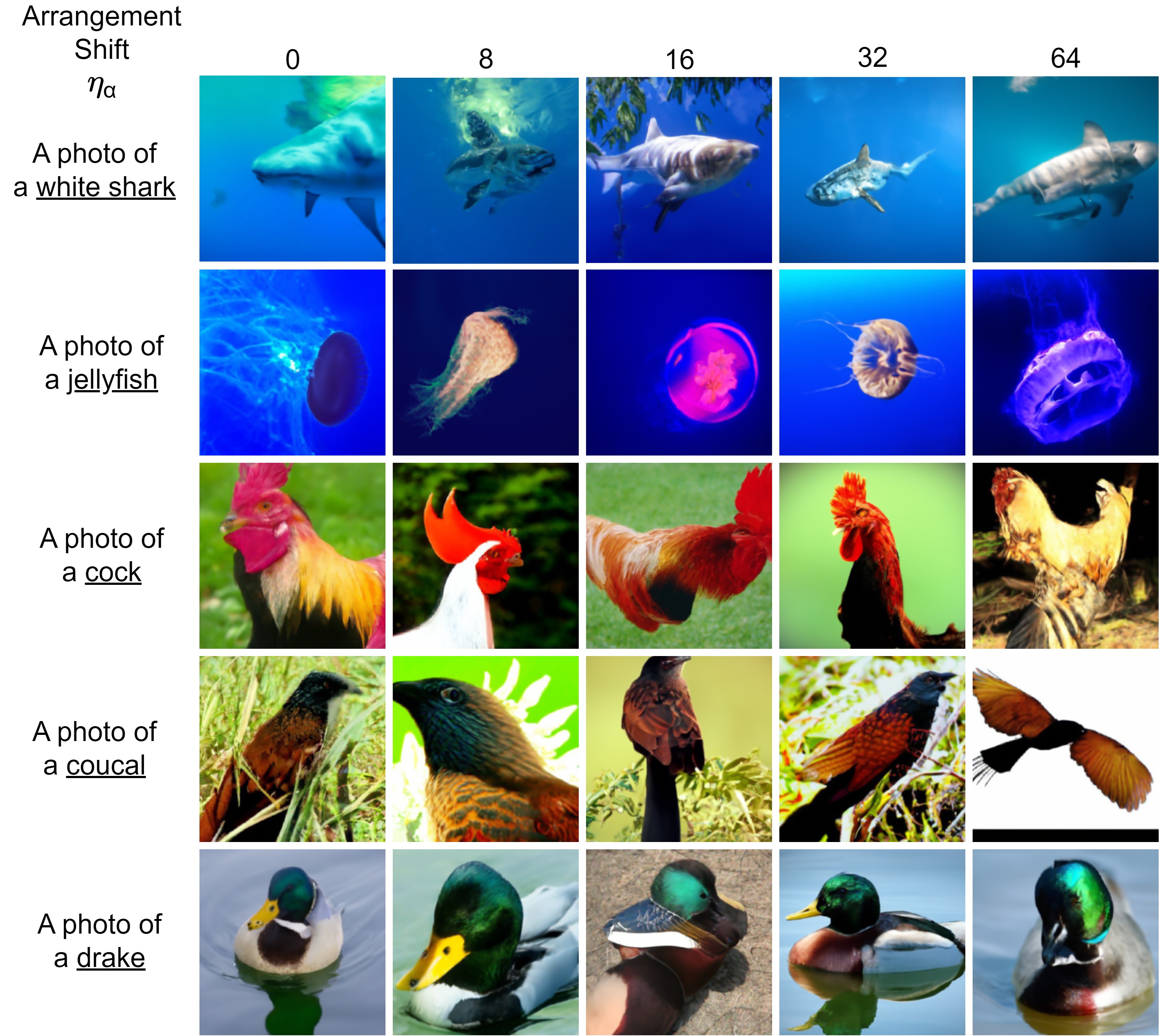}
    \caption{
    \textbf{Visual Inspection of image generated with \AS~ by \Glide.}}
    \label{fig:appendix_glide_as}
\end{figure*}

%% file: _arxiv.bbl
\begin{thebibliography}{40}
\providecommand{\natexlab}[1]{#1}
\providecommand{\url}[1]{\texttt{#1}}
\expandafter\ifx\csname urlstyle\endcsname\relax
  \providecommand{\doi}[1]{doi: #1}\else
  \providecommand{\doi}{doi: \begingroup \urlstyle{rm}\Url}\fi

\bibitem[Bao et~al.(2022)Bao, Li, Zhu, and Zhang]{bao2022analyticdpm}
Fan Bao, Chongxuan Li, Jun Zhu, and Bo~Zhang.
\newblock Analytic-{DPM}: an analytic estimate of the optimal reverse variance
  in diffusion probabilistic models.
\newblock In \emph{International Conference on Learning Representations}, 2022.
\newblock URL \url{https://openreview.net/forum?id=0xiJLKH-ufZ}.

\bibitem[Chefer et~al.(2023)Chefer, Alaluf, Vinker, Wolf, and
  Cohen-Or]{chefer2023attend}
Hila Chefer, Yuval Alaluf, Yael Vinker, Lior Wolf, and Daniel Cohen-Or.
\newblock Attend-and-excite: Attention-based semantic guidance for
  text-to-image diffusion models.
\newblock \emph{ACM Transactions on Graphics (TOG)}, 42\penalty0 (4):\penalty0
  1--10, 2023.

\bibitem[Daras et~al.(2023)Daras, Dagan, Dimakis, and
  Daskalakis]{daras2023consistent}
Giannis Daras, Yuval Dagan, Alexandros~G Dimakis, and Constantinos Daskalakis.
\newblock Consistent diffusion models: Mitigating sampling drift by learning to
  be consistent.
\newblock \emph{arXiv preprint arXiv:2302.09057}, 2023.

\bibitem[Dhariwal \& Nichol(2021)Dhariwal and Nichol]{dhariwal2021diffusion}
Prafulla Dhariwal and Alexander Nichol.
\newblock Diffusion models beat gans on image synthesis.
\newblock \emph{Advances in neural information processing systems},
  34:\penalty0 8780--8794, 2021.

\bibitem[Feng et~al.(2022)Feng, He, Fu, Jampani, Akula, Narayana, Basu, Wang,
  and Wang]{feng2022training}
Weixi Feng, Xuehai He, Tsu-Jui Fu, Varun Jampani, Arjun Akula, Pradyumna
  Narayana, Sugato Basu, Xin~Eric Wang, and William~Yang Wang.
\newblock Training-free structured diffusion guidance for compositional
  text-to-image synthesis.
\newblock \emph{arXiv preprint arXiv:2212.05032}, 2022.

\bibitem[Ge et~al.(2023)Ge, Nah, Liu, Poon, Tao, Catanzaro, Jacobs, Huang, Liu,
  and Balaji]{ge2023preserve}
Songwei Ge, Seungjun Nah, Guilin Liu, Tyler Poon, Andrew Tao, Bryan Catanzaro,
  David Jacobs, Jia-Bin Huang, Ming-Yu Liu, and Yogesh Balaji.
\newblock Preserve your own correlation: A noise prior for video diffusion
  models.
\newblock In \emph{Proceedings of the IEEE/CVF International Conference on
  Computer Vision}, pp.\  22930--22941, 2023.

\bibitem[Goodfellow et~al.(2014)Goodfellow, Pouget-Abadie, Mirza, Xu,
  Warde-Farley, Ozair, Courville, and Bengio]{goodfellow2014generative}
Ian Goodfellow, Jean Pouget-Abadie, Mehdi Mirza, Bing Xu, David Warde-Farley,
  Sherjil Ozair, Aaron Courville, and Yoshua Bengio.
\newblock Generative adversarial nets.
\newblock \emph{Advances in neural information processing systems}, 27, 2014.

\bibitem[Ho \& Salimans(2022)Ho and Salimans]{ho2022classifier}
Jonathan Ho and Tim Salimans.
\newblock Classifier-free diffusion guidance.
\newblock \emph{arXiv preprint arXiv:2207.12598}, 2022.

\bibitem[Ho et~al.(2020)Ho, Jain, and Abbeel]{ho2020denoising}
Jonathan Ho, Ajay Jain, and Pieter Abbeel.
\newblock Denoising diffusion probabilistic models.
\newblock \emph{Advances in neural information processing systems},
  33:\penalty0 6840--6851, 2020.

\bibitem[Ho et~al.(2022)Ho, Saharia, Chan, Fleet, Norouzi, and
  Salimans]{ho2022cascaded}
Jonathan Ho, Chitwan Saharia, William Chan, David~J Fleet, Mohammad Norouzi,
  and Tim Salimans.
\newblock Cascaded diffusion models for high fidelity image generation.
\newblock \emph{The Journal of Machine Learning Research}, 23\penalty0
  (1):\penalty0 2249--2281, 2022.

\bibitem[Huang et~al.(2022{\natexlab{a}})Huang, Lam, Wang, Su, Yu, Ren, and
  Zhao]{ijcai2022p577}
Rongjie Huang, Max W.~Y. Lam, Jun Wang, Dan Su, Dong Yu, Yi~Ren, and Zhou Zhao.
\newblock Fastdiff: A fast conditional diffusion model for high-quality speech
  synthesis.
\newblock In Lud~De Raedt (ed.), \emph{Proceedings of the Thirty-First
  International Joint Conference on Artificial Intelligence, {IJCAI-22}}, pp.\
  4157--4163. International Joint Conferences on Artificial Intelligence
  Organization, 7 2022{\natexlab{a}}.
\newblock \doi{10.24963/ijcai.2022/577}.
\newblock URL \url{https://doi.org/10.24963/ijcai.2022/577}.
\newblock Main Track.

\bibitem[Huang et~al.(2022{\natexlab{b}})Huang, Zhao, Liu, Liu, Cui, and
  Ren]{huang2022prodiff}
Rongjie Huang, Zhou Zhao, Huadai Liu, Jinglin Liu, Chenye Cui, and Yi~Ren.
\newblock Prodiff: Progressive fast diffusion model for high-quality
  text-to-speech.
\newblock In \emph{Proceedings of the 30th ACM International Conference on
  Multimedia}, pp.\  2595--2605, 2022{\natexlab{b}}.

\bibitem[Jeong et~al.(2023)Jeong, Ryoo, Lee, Seo, Byeon, Kim, and
  Kim]{jeong2023power}
Yujin Jeong, Wonjeong Ryoo, Seunghyun Lee, Dabin Seo, Wonmin Byeon, Sangpil
  Kim, and Jinkyu Kim.
\newblock The power of sound (tpos): Audio reactive video generation with
  stable diffusion.
\newblock In \emph{Proceedings of the IEEE/CVF International Conference on
  Computer Vision}, pp.\  7822--7832, 2023.

\bibitem[Kim et~al.(2022)Kim, Kim, and Yoon]{kim2022guided}
Heeseung Kim, Sungwon Kim, and Sungroh Yoon.
\newblock Guided-tts: A diffusion model for text-to-speech via classifier
  guidance.
\newblock In \emph{International Conference on Machine Learning}, pp.\
  11119--11133. PMLR, 2022.

\bibitem[Kong et~al.(2021)Kong, Ping, Huang, Zhao, and
  Catanzaro]{kong2021diffwave}
Zhifeng Kong, Wei Ping, Jiaji Huang, Kexin Zhao, and Bryan Catanzaro.
\newblock Diffwave: A versatile diffusion model for audio synthesis.
\newblock In \emph{International Conference on Learning Representations}, 2021.
\newblock URL \url{https://openreview.net/forum?id=a-xFK8Ymz5J}.

\bibitem[Li et~al.(2023{\natexlab{a}})Li, Qu, Sun, and
  Moens]{li2023alleviating}
Mingxiao Li, Tingyu Qu, Wei Sun, and Marie-Francine Moens.
\newblock Alleviating exposure bias in diffusion models through sampling with
  shifted time steps.
\newblock \emph{arXiv preprint arXiv:2305.15583}, 2023{\natexlab{a}}.

\bibitem[Li et~al.(2022)Li, Thickstun, Gulrajani, Liang, and
  Hashimoto]{li2022diffusion}
Xiang Li, John Thickstun, Ishaan Gulrajani, Percy~S Liang, and Tatsunori~B
  Hashimoto.
\newblock Diffusion-lm improves controllable text generation.
\newblock \emph{Advances in Neural Information Processing Systems},
  35:\penalty0 4328--4343, 2022.

\bibitem[Li et~al.(2023{\natexlab{b}})Li, Qian, and van~der
  Schaar]{li2023diffusion}
Yangming Li, Zhaozhi Qian, and Mihaela van~der Schaar.
\newblock Do diffusion models suffer error propagation? theoretical analysis
  and consistency regularization.
\newblock \emph{arXiv preprint arXiv:2308.05021}, 2023{\natexlab{b}}.

\bibitem[Liu et~al.(2022)Liu, Li, Du, Torralba, and
  Tenenbaum]{liu2022compositional}
Nan Liu, Shuang Li, Yilun Du, Antonio Torralba, and Joshua~B Tenenbaum.
\newblock Compositional visual generation with composable diffusion models.
\newblock In \emph{Computer Vision--ECCV 2022: 17th European Conference, Tel
  Aviv, Israel, October 23--27, 2022, Proceedings, Part XVII}, pp.\  423--439.
  Springer, 2022.

\bibitem[Lugmayr et~al.(2022)Lugmayr, Danelljan, Romero, Yu, Timofte, and
  Van~Gool]{lugmayr2022repaint}
Andreas Lugmayr, Martin Danelljan, Andres Romero, Fisher Yu, Radu Timofte, and
  Luc Van~Gool.
\newblock Repaint: Inpainting using denoising diffusion probabilistic models.
\newblock In \emph{Proceedings of the IEEE/CVF Conference on Computer Vision
  and Pattern Recognition}, pp.\  11461--11471, 2022.

\bibitem[Mao et~al.(2023)Mao, Wang, and Aizawa]{mao2023guided}
Jiafeng Mao, Xueting Wang, and Kiyoharu Aizawa.
\newblock Guided image synthesis via initial image editing in diffusion model.
\newblock \emph{arXiv preprint arXiv:2305.03382}, 2023.

\bibitem[Mirza \& Osindero(2014)Mirza and Osindero]{mirza2014conditional}
Mehdi Mirza and Simon Osindero.
\newblock Conditional generative adversarial nets.
\newblock \emph{arXiv preprint arXiv:1411.1784}, 2014.

\bibitem[Nichol et~al.(2022)Nichol, Dhariwal, Ramesh, Shyam, Mishkin, McGrew,
  Sutskever, and Chen]{DBLP:conf/icml/NicholDRSMMSC22}
Alexander~Quinn Nichol, Prafulla Dhariwal, Aditya Ramesh, Pranav Shyam, Pamela
  Mishkin, Bob McGrew, Ilya Sutskever, and Mark Chen.
\newblock {GLIDE:} towards photorealistic image generation and editing with
  text-guided diffusion models.
\newblock In Kamalika Chaudhuri, Stefanie Jegelka, Le~Song, Csaba
  Szepesv{\'{a}}ri, Gang Niu, and Sivan Sabato (eds.), \emph{International
  Conference on Machine Learning, {ICML} 2022, 17-23 July 2022, Baltimore,
  Maryland, {USA}}, volume 162 of \emph{Proceedings of Machine Learning
  Research}, pp.\  16784--16804. {PMLR}, 2022.
\newblock URL \url{https://proceedings.mlr.press/v162/nichol22a.html}.

\bibitem[Ning et~al.(2023)Ning, Sangineto, Porrello, Calderara, and
  Cucchiara]{ning2023input}
Mang Ning, Enver Sangineto, Angelo Porrello, Simone Calderara, and Rita
  Cucchiara.
\newblock Input perturbation reduces exposure bias in diffusion models.
\newblock \emph{arXiv preprint arXiv:2301.11706}, 2023.

\bibitem[Radford et~al.(2021)Radford, Kim, Hallacy, Ramesh, Goh, Agarwal,
  Sastry, Askell, Mishkin, Clark, et~al.]{radford2021learning}
Alec Radford, Jong~Wook Kim, Chris Hallacy, Aditya Ramesh, Gabriel Goh,
  Sandhini Agarwal, Girish Sastry, Amanda Askell, Pamela Mishkin, Jack Clark,
  et~al.
\newblock Learning transferable visual models from natural language
  supervision.
\newblock In \emph{International conference on machine learning}, pp.\
  8748--8763. PMLR, 2021.

\bibitem[Ramesh et~al.(2022)Ramesh, Dhariwal, Nichol, Chu, and
  Chen]{ramesh2022hierarchical}
Aditya Ramesh, Prafulla Dhariwal, Alex Nichol, Casey Chu, and Mark Chen.
\newblock Hierarchical text-conditional image generation with clip latents.
\newblock \emph{arXiv preprint arXiv:2204.06125}, 1\penalty0 (2):\penalty0 3,
  2022.

\bibitem[Rombach et~al.(2022)Rombach, Blattmann, Lorenz, Esser, and
  Ommer]{stable_dif}
Robin Rombach, Andreas Blattmann, Dominik Lorenz, Patrick Esser, and Bj{\"o}rn
  Ommer.
\newblock High-resolution image synthesis with latent diffusion models.
\newblock In \emph{Proceedings of the IEEE/CVF conference on computer vision
  and pattern recognition}, pp.\  10684--10695, 2022.

\bibitem[Saharia et~al.(2022)Saharia, Chan, Saxena, Li, Whang, Denton,
  Ghasemipour, Gontijo~Lopes, Karagol~Ayan, Salimans,
  et~al.]{saharia2022photorealistic}
Chitwan Saharia, William Chan, Saurabh Saxena, Lala Li, Jay Whang, Emily~L
  Denton, Kamyar Ghasemipour, Raphael Gontijo~Lopes, Burcu Karagol~Ayan, Tim
  Salimans, et~al.
\newblock Photorealistic text-to-image diffusion models with deep language
  understanding.
\newblock \emph{Advances in Neural Information Processing Systems},
  35:\penalty0 36479--36494, 2022.

\bibitem[Salimans \& Ho(2022)Salimans and Ho]{salimans2022progressive}
Tim Salimans and Jonathan Ho.
\newblock Progressive distillation for fast sampling of diffusion models.
\newblock In \emph{International Conference on Learning Representations}, 2022.
\newblock URL \url{https://openreview.net/forum?id=TIdIXIpzhoI}.

\bibitem[Samuel et~al.(2023{\natexlab{a}})Samuel, Ben-Ari, Darshan, Maron, and
  Chechik]{samuel2023norm}
Dvir Samuel, Rami Ben-Ari, Nir Darshan, Haggai Maron, and Gal Chechik.
\newblock Norm-guided latent space exploration for text-to-image generation.
\newblock \emph{arXiv preprint arXiv:2306.08687}, 2023{\natexlab{a}}.

\bibitem[Samuel et~al.(2023{\natexlab{b}})Samuel, Ben-Ari, Raviv, Darshan, and
  Chechik]{samuel2023all}
Dvir Samuel, Rami Ben-Ari, Simon Raviv, Nir Darshan, and Gal Chechik.
\newblock It is all about where you start: Text-to-image generation with seed
  selection.
\newblock \emph{arXiv preprint arXiv:2304.14530}, 2023{\natexlab{b}}.

\bibitem[SHEKHAR(2021)]{ImageNet100}
AMBESH SHEKHAR.
\newblock {{ImageNet100}}.
\newblock https://www.kaggle.com/datasets/ambityga/imagenet100, 2021.

\bibitem[Singh et~al.(2022{\natexlab{a}})Singh, Gustafson, Adcock,
  de~Freitas~Reis, Gedik, Kosaraju, Mahajan, Girshick, Doll{\'a}r, and Van
  Der~Maaten]{singh2022revisiting}
Mannat Singh, Laura Gustafson, Aaron Adcock, Vinicius de~Freitas~Reis, Bugra
  Gedik, Raj~Prateek Kosaraju, Dhruv Mahajan, Ross Girshick, Piotr Doll{\'a}r,
  and Laurens Van Der~Maaten.
\newblock Revisiting weakly supervised pre-training of visual perception
  models.
\newblock In \emph{Proceedings of the IEEE/CVF Conference on Computer Vision
  and Pattern Recognition}, pp.\  804--814, 2022{\natexlab{a}}.

\bibitem[Singh et~al.(2022{\natexlab{b}})Singh, Jandial, Chopra, Ramesh,
  Krishnamurthy, and Balasubramanian]{singh2022conditioning}
Vedant Singh, Surgan Jandial, Ayush Chopra, Siddharth Ramesh, Balaji
  Krishnamurthy, and Vineeth~N Balasubramanian.
\newblock On conditioning the input noise for controlled image generation with
  diffusion models.
\newblock \emph{arXiv preprint arXiv:2205.03859}, 2022{\natexlab{b}}.

\bibitem[Sohl-Dickstein et~al.(2015)Sohl-Dickstein, Weiss, Maheswaranathan, and
  Ganguli]{diff_2015}
Jascha Sohl-Dickstein, Eric Weiss, Niru Maheswaranathan, and Surya Ganguli.
\newblock Deep unsupervised learning using nonequilibrium thermodynamics.
\newblock In Francis Bach and David Blei (eds.), \emph{Proceedings of the 32nd
  International Conference on Machine Learning}, volume~37 of \emph{Proceedings
  of Machine Learning Research}, pp.\  2256--2265, Lille, France, 07--09 Jul
  2015. PMLR.
\newblock URL \url{https://proceedings.mlr.press/v37/sohl-dickstein15.html}.

\bibitem[Sohn et~al.(2015)Sohn, Lee, and Yan]{sohn2015learning}
Kihyuk Sohn, Honglak Lee, and Xinchen Yan.
\newblock Learning structured output representation using deep conditional
  generative models.
\newblock \emph{Advances in neural information processing systems}, 28, 2015.

\bibitem[Song et~al.(2020)Song, Meng, and Ermon]{song2020denoising}
Jiaming Song, Chenlin Meng, and Stefano Ermon.
\newblock Denoising diffusion implicit models.
\newblock \emph{arXiv preprint arXiv:2010.02502}, 2020.

\bibitem[Wu \& De~la Torre(2022)Wu and De~la Torre]{wu2022making}
Chen~Henry Wu and Fernando De~la Torre.
\newblock Making text-to-image diffusion models zero-shot image-to-image
  editors by inferring" random seeds".
\newblock In \emph{NeurIPS 2022 Workshop on Score-Based Methods}, 2022.

\bibitem[Xu et~al.(2022)Xu, Yu, Song, Shi, Ermon, and Tang]{xu2022geodiff}
Minkai Xu, Lantao Yu, Yang Song, Chence Shi, Stefano Ermon, and Jian Tang.
\newblock Geodiff: A geometric diffusion model for molecular conformation
  generation.
\newblock In \emph{International Conference on Learning Representations}, 2022.
\newblock URL \url{https://openreview.net/forum?id=PzcvxEMzvQC}.

\bibitem[Zhao et~al.(2019)Zhao, Meng, Yin, and Sigal]{zhao2019image}
Bo~Zhao, Lili Meng, Weidong Yin, and Leonid Sigal.
\newblock Image generation from layout.
\newblock In \emph{Proceedings of the IEEE/CVF Conference on Computer Vision
  and Pattern Recognition}, pp.\  8584--8593, 2019.

\end{thebibliography}
